\documentclass[journal]{IEEEtran}
\let\labelindent\relax

\usepackage[cmex10]{amsmath}
\usepackage{graphicx, algorithm, float}
\usepackage{amssymb,mathtools} 
\usepackage{enumitem}
\usepackage{color}
\usepackage{subfig}
\usepackage[noend]{algpseudocode}
\usepackage{fixltx2e}
\usepackage{adjustbox}
\usepackage{pbox}

\newcommand{\etal}{\textit{et al}.}
\newcommand{\ie}{\textit{i}.\textit{e}.,\,}
\newcommand{\eg}{\textit{e}.\textit{g}.,\,}
\def\NoNumber#1{{\def\alglinenumber##1{}\State #1}\addtocounter{ALG@line}{-1}}
\DeclareMathOperator*{\argmin}{\arg\!\min}
\DeclarePairedDelimiter\norm{\lVert}{\rVert}
\newcommand\Tstrut{\rule{0pt}{2.6ex}}         % = `top' strut
\newcommand\Bstrut{\rule[-0.9ex]{0pt}{0pt}}   % = `bottom' strut
\begin{document}

\title{Object Classification with Joint Projection and Low-rank Dictionary Learning}

\author{Homa~Foroughi, Nilanjan~Ray and~Hong~Zhang 
\thanks{H. Foroughi is with the Department
of Computing Science, University of Alberta, Edmonton,
AB, T6G2E8, Canada, e-mail: homa@ualberta.ca.}% <-this % stops a space
\thanks{N. Ray and H. Zhnag are with Department of Computing Science, University of Alberta.}}

%\markboth{IEEE Transactions on Image Processing}%

\maketitle
%|||||||||||||||||||||||||||||||||||||||||||||||||||||||||||||||||||||||||||||||||||||||||||||||||||||||||
%||-----------------------------------------------------------------------------------------------------||
%||-----------------------------------------------------------------------------------------------------||
%||----------------------------------- Abstract --------------------------------------------------------||
%||-----------------------------------------------------------------------------------------------------||
%||-----------------------------------------------------------------------------------------------------||
%|||||||||||||||||||||||||||||||||||||||||||||||||||||||||||||||||||||||||||||||||||||||||||||||||||||||||
\begin{abstract}
For an object classification system, the most critical obstacles towards real-world applications are often caused by large intra-class variability, arising from different lightings, occlusion and corruption, in limited sample sets. Most methods in the literature would fail when the training samples are heavily occluded, corrupted or have significant illumination or viewpoint variations. Besides, most of the existing methods and especially deep learning-based methods, need large training sets to achieve a satisfactory recognition performance. Although using the pre-trained network on a generic large-scale dataset and fine-tune it to the small-sized target dataset is a widely used technique, this would not help when the content of base and target datasets are very different.
To address these issues, we propose a joint projection and low-rank dictionary learning method using dual graph constraints (JP-LRDL). The proposed joint learning method would enable us to learn the features on top of which dictionaries can be better learned, from the data with large intra-class variability. Specifically, a structured class-specific dictionary is learned and the discrimination is further improved by imposing a graph constraint on the coding coefficients, that maximizes the intra-class compactness and inter-class separability. We also enforce low-rank and structural incoherence constraints on sub-dictionaries to make them more compact and robust to variations and outliers and reduce the redundancy among them, respectively. To preserve the intrinsic structure of data and penalize unfavourable relationship among training samples simultaneously, we introduce a projection graph into the framework, which significantly enhances the discriminative ability of the projection matrix and makes the method robust to small-sized and high-dimensional datasets. Experimental results on several benchmark datasets verify the superior performance of our method for object classification of small datasets, which include considerable amount of different kinds of variation.
\end{abstract}
%|||||||||||||||||||||||||||||||||||||||||||||||||||||||||||||||||||||||||||||||||||||||||||||||||||||||||
%||-----------------------------------------------------------------------------------------------------||
%||-----------------------------------------------------------------------------------------------------||
%||----------------------------------- Introduction ----------------------------------------------------||
%||-----------------------------------------------------------------------------------------------------||
%||-----------------------------------------------------------------------------------------------------||
%|||||||||||||||||||||||||||||||||||||||||||||||||||||||||||||||||||||||||||||||||||||||||||||||||||||||||
\section{Introduction}
\label{sec:intro}
Image classification based on visual content is a very challenging task, mainly because there is usually large amount of intra-class variability, arising from illumination and viewpoint variations, occlusion and corruption~\cite{PCA-Net}. Numerous efforts have been made to counter the intra-class variability by manually designing low-level features for classification tasks. Representative examples are Gabor features and LBP~\cite{LBP} for texture and face classification, and SIFT~\cite{SIFT} and HOG~\cite{HOG} features for object recognition. Although the hand-crafted low-level features achieve great success for some controlled scenarios, designing effective features for new data and tasks usually requires new domain knowledge since most hand-crafted features cannot be simply adopted to new conditions. Learning features from data itself instead of manually designing features is considered a plausible way to overcome the limitation of low-level features~\cite{PCA-Net}, and successful examples of such methods are dictionary learning and deep learning. 

The idea of deep learning is to discover multiple levels of representation, with the hope that higher level features represent more abstract semantics of the data. Such abstract representations learned from a deep network are expected to provide more invariance to intra-class variability, if we train the deep model using a large amount of training samples~\cite{PCA-Net}. One key ingredient for this success is the use of convolutional architectures. A convolutional neural network (CNN) architecture consists of multiple trainable stages stacked on top of each other, followed by a supervised classifier. In practice, many computer vision applications are faced with the problem of small training sets, and transfer learning can be a powerful tool to enable training the target network in such cases. The usual approach is to replace and retrain the classifier on top of the CNN on the target dataset, and also fine-tune the weights of the pretrained network by continuing the backpropagation. However, the effectiveness of feature transfer is declined when the base and target tasks become less similar~\cite{Transferable}. Besides, when the target dataset is small, complex models like CNNs, tend to overfit the data easily~\cite{Transferable}. It could be even more complicated in classification tasks such as face recognition, which the intra-class variability is often greater than the inter-class variability due to pose, expression and illumination changes and occlusion.

In contrast, the recent variations of dictionary learning (DL) methods have demonstrated great success in image classification tasks on both small and large intra-class variation datasets. The last few years have witnessed fast development on DL methods under sparse representation theory, accordding to which, signals can be well-approximated by linear combination of a few columns of some appropriate basis or dictionary~\cite{SRC}. The dictionary, which should faithfully and discriminatively represent the encoded signal, plays an important role in the success of sparse representation and it has been shown that learned dictionaries significantly outperform pre-defined ones such as Wavelets~\cite{FDDL}. 

Although conventional DL methods perform well for different classification and recognition tasks~\cite{FDDL}, their performance dramatically deteriorates when the training data are contaminated heavily because of occlusion, lighting/viewpoint variations or corruption. In the recent years, low-rank (LR) matrix recovery, which efficiently removes noise from corrupted observations, has been successfully applied to a variety of computer vision applications, such as subspace clustering~\cite{LR-Subspace}, background subtraction~\cite{BS-LR} and image classification~\cite{BMVC-Homa}. Accordingly, some DL methods have been proposed by integrating rank minimization into sparse representation framework and achieved impressive results, especially when large noise exists~\cite{D2L2R2},~\cite{Structured-LR-DL}. 

Moreover, in many areas of computer vision, data are characterized by high dimensional feature vectors; however, dealing with high-dimensional data is challenging for many tasks such as DL. High-dimensional data are not only inefficient and computationally intensive, but the sheer number of dimensions often masks the discriminative signal embedded in the data~\cite{LGE-KSVD}. As a solution, a dimensionality reduction (DR) technique is usually performed first on the training samples, and the dimensionality reduced data are then used as the input of DL. However, recent studies reveal that the pre-learned projection matrices neither fully promote the underlying sparse structure of data~\cite{SE}, nor preserve the best features for DL~\cite{JDDRDL}. Intuitively, the DR and DL processes should be jointly conducted for a more effective classification. 

Only a few works have discussed the idea of jointly learning the projection of training samples and dictionary, and all reported more competitive performance than conventional DL methods. Despite the successes, most of the existing joint DR-DL methods cannot handle noisy (occluded/corrupted) and large intra-class observations. On the other hand, low-rank DL methods can cope well with noisy data, but cannot select the best features on top of which dictionaries can be better learned, due to separated DR process. In this paper, we explore the DL, LR and DR spaces simultaneously and propose an object classification method for noisy and large intra-class variation datasets, which have small-sized training set and may have high-dimensional feature vectors. To the best of our knowledge, this is the first proposed method that can handle all these issues simultaneously.

\begin{figure}[t]
\centering
\includegraphics[width=7cm,keepaspectratio]{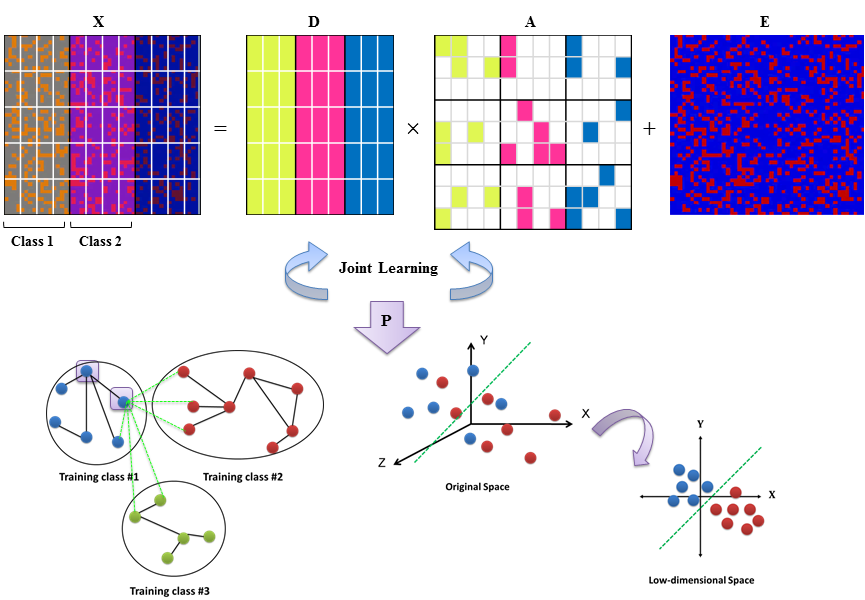}
\caption{JP-LRDL framework}
\label{fig:Diagram}
\vspace{-1.5em}
\end{figure}

To this end, we propose a novel framework called \textit{joint projection and low-rank dictionary learning using dual graph constraints} (JP-LRDL). The basic idea of JP-LRDL is illustrated in Figure~\ref{fig:Diagram}. The algorithm learns a discriminative structured dictionary in the reduced space, whose atoms have correspondence to the class labels and a graph constraint is imposed on the coding vectors to further enhance class discrimination. The \textit{coefficient} graph makes the coding coefficients within the same class to be similar and the coefficients among different classes to be dissimilar. JP-LRDL specially introduces low-rank and incoherence promoting constraints on sub-dictionaries to make them more compact and robust to variations and encourage them to be as independent as possible, respectively. Simultaneously, we consider optimizing the input feature space by jointly learning a feature projection matrix. In particular, another graph is built on training data to explore intrinsic geometric structure of data. The \textit{projection} graph enables us to preserve the desirable relationship among training samples and penalize the unfavorable relationships simultaneously. This joint framework empowers our algorithm with several important advantages: (1) Ability to handle large intra-class variation observations, (2) Promoting the discriminative ability of the learned projection and dictionary, that enables us to deal with small-sized datasets, (3) Learning in the reduced dimensions with lower computational complexity, and (4) Maintaining both global and local structure of data. Extensive experimental results validate the effectiveness of our method and its applicability to image classification task, especially for noisy observations. 

The remainder of the paper is organized as follows. Section~\ref{sec:rel} briefly reviews some related work. Section~\ref{sec:proposed} presents the proposed JP-LRDL method. The optimization algorithms are described in Section~\ref{sec:opt}. We discuss the time complexity and convergence analysis in Section~\ref{sec:comp}. The classification scheme is then explained in Section~\ref{sec:classification}. Section~\ref{sec:experiment} shows experimental results and we draw conclusions in Section~\ref{sec:conclusion}.
%&&&&&&&&&&&&&&&&&&&&&&&&&&&&&&&&&&&&&&&&&&&&&&&&&&&&&&&&&&&&&&&&&&&&&&&&&&&&&&&&&&&&&&&&&&&&&&&&&&&&&&&&
%------------------------------------------- Diagram ----------------------------------------------------
%-------------------------------------------------------------------------------------------------------
%&&&&&&&&&&&&&&&&&&&&&&&&&&&&&&&&&&&&&&&&&&&&&&&&&&&&&&&&&&&&&&&&&&&&&&&&&&&&&&&&&&&&&&&&&&&&&&&&&&&&&&&&
%|||||||||||||||||||||||||||||||||||||||||||||||||||||||||||||||||||||||||||||||||||||||||||||||||||||||||
%||-----------------------------------------------------------------------------------------------------||
%||-----------------------------------------------------------------------------------------------------||
%||----------------------------------- Related Work ----------------------------------------------------||
%||-----------------------------------------------------------------------------------------------------||
%||-----------------------------------------------------------------------------------------------------||
%|||||||||||||||||||||||||||||||||||||||||||||||||||||||||||||||||||||||||||||||||||||||||||||||||||||||||
\section{Related Work}
\label{sec:rel}
The goal of DL is to learn a dictionary which can yield sparse representation of training samples. Supervised DL methods have been proposed to promote the discriminative power of the learned dictionary by exploiting the information among classes. Related literature include the discriminative K-SVD~\cite{DKSVD} and label-consistent K-SVD~\cite{LC-KSVD}. For stronger discrimination, Yang \etal~\cite{FDDL} proposed a Fisher discrimination dictionary learning (FDDL) method by imposing Fisher discrimination criterion on the coding vectors. More recently,~\cite{Latent-DL} proposed a latent DL method by jointly learning a latent matrix to adaptively build the relationship between dictionary atoms and class labels.

The recent advances in LR learning have shown excellent performance for handling large noise and variation. Accordingly, some DL approaches have been proposed recently based on LR approximation to improve the classification performance with noise. Ma \etal~\cite{DLRD-SR} integrated rank minimization into sparse representation by introducing LR constraints on sub-dictionaries for each class. To make the dictionary more discerning, Li \etal~\cite{D2L2R2} proposed a discriminative DL method, called D\textsuperscript{2}L\textsuperscript{2}R\textsuperscript{2}, which adopts Fisher discrimination and meantime imposes a LR constraint on sub-dictionaries to make them robust to variations and achieves impressive results especially when corruption existed. Zhang \etal~\cite{Structured-LR-DL} proposed a discriminative, structured low-rank DL method to explore the global structure among all training samples. A code regularization term with block-diagonal structure is incorporated to learn discriminative dictionary, which regularizes the same class images to have the same representation.

As mentioned earlier, to deal with high-dimensional feature vectors in DL process, DR is usually performed first on the training samples by either PCA~\cite{PCA} or Ranodm Projection~\cite{RP} techniques and then, the dimensionality reduced data are used for DL process. However, recent studies reveal that the joint DR and DL frameworks, make the learned projection and dictionary a better fit for each other, so a more accurate classification can be obtained. Nevertheless, only a few works have discussed the idea of jointly learning the projection of training samples and dictionary.

\cite{Simul-SRC} presented a simultaneous projection and DL method using a carefully designed sigmoid reconstruction error. The data is projected to an orthogonal space where the intra- and inter-class reconstruction errors are minimized and maximized, respectively for making the projected space discriminative. However,~\cite{Invariance} showed that the dictionary learned in the projected space is not more discriminative than the one learned in the original space. JDDLDR method~\cite{JDDRDL} jointly learns a DR matrix and a discriminative dictionary and achieves promising results for face recognition. The discrimination is enforced by a Fisher-like constraint on the coding coefficients, but the projection matrix is learned without any discrimination constraints. Nguyen \etal~\cite{SE} proposed a joint DR and sparse learning framework by emphasizing on preserving the sparse structure of data. Their method, known as sparse embedding (SE), can be extended to a non-linear version via kernel tricks and also adopts a novel classification schema leading to great performance. However, it fails to consider the discrimination power among the separately learned class-specific dictionaries, such that it is not guaranteed to produce improved classification performance~\cite{Problem-SE}. 
Ptucha \etal~\cite{LGE-KSVD} integrated manifold-based DR and sparse representation within a single framework and presented a variant of the K-SVD algorithm by exploiting a linear extension of graph embedding (LGE). The LGE concept is further leveraged to modify the K-SVD algorithm for co-optimizing a small, yet over-complete dictionary, the projection matrix and the coefficients. Yang \etal~\cite{Simul-DL} learns a DR projection matrix and a class-specific dictionary simultaneously and exploits both representation residuals and coefficients for the classification purpose. Most recently, Liu \etal~\cite{JNPDL} proposed a joint non-negative projection and DL method. The discrimination is achieved by imposing graph constraints on both projection and coding coefficients that maximizes the intra-class compactness and inter-class separability.

Inspired by the related works, we aim at proposing an object classification method based on joint DR and low-rank DL framework. By imposing appropriate constraints into the objective function, the proposed JP-LRDL method would be robust to large intra-class variability and small-sized datasets. Besides, the joint framework would enable us to learn features from high-dimensional data, on top of which dictionaries can be better learned for a more accurate classification.
%|||||||||||||||||||||||||||||||||||||||||||||||||||||||||||||||||||||||||||||||||||||||||||||||||||||||||
%||-----------------------------------------------------------------------------------------------------||
%||-----------------------------------------------------------------------------------------------------||
%||----------------------------------- Proposed Method -------------------------------------------------||
%||-----------------------------------------------------------------------------------------------------||
%||-----------------------------------------------------------------------------------------------------||
%|||||||||||||||||||||||||||||||||||||||||||||||||||||||||||||||||||||||||||||||||||||||||||||||||||||||||
\section{The Proposed JP-LRDL Framework}
\label{sec:proposed}
We aim to learn a discriminative dictionary and a robust projection matrix simultaneously, using LR regularization and dual graph constraints. 
Let $X$ be a set of $m$-dimensional training samples, \ie $X=\{ X_1, X_2, \dots , X_K \}$, where $X_i$ denotes the training samples from class $i$ and $K$ is the number of classes. The class-specific dictionary is denoted by $D=\{ D_1, D_2, \dots , D_K \}$, where $D_i$ is the sub-dictionary associated with class $i$. 
We also want to learn the projection matrix $P \in R^{m \times d} \, (d<m)$, that projects data into a low-dimensional space. Denote by $A$ the sparse representation matrix of the dimensionality reduced data $P^T X$ over dictionary $D$, \ie, $P^T X \approx DA$. We can write $A$ as $A=\{ A_1, A_2, \dots , A_K \}$, where $A_i$ is the representation of $P^T X_i$ over $D$. Therefore, we propose JP-LRDL optimization model:
\begin{gather}
J_{(P,D,A)} = \argmin_{P,D,A} \Big\{ R(P,D,A) + \lambda_{1} \, \norm[\big]{A}_{1} + \lambda_{2} \, G(A)   
\label{eq1} \\ \notag
+ \, \lambda_{3} \sum\limits_{i}  \norm[\big]{D_i}_{*}  +  \delta \, G(P) \Big\} \quad s.t. \quad P^TP=I
\end{gather}
where $R(P,D,A)$ is the discriminative reconstruction error, $\norm{A}_{1}$ denotes the $l_1$-regularization on coding coefficients, $G(A)$ is the graph-based coding coefficients, $\norm{D_i}_{*}$ is the nuclear norm of sub-dictionary $D_i \,$, $G(P)$ represents the graph-based projection and $\lambda_{1}, \lambda_{2}, \lambda_{3}, \delta$ are scalar parameters. %We will discuss these terms in details in the following section.
%---------------------------------------------------------------------------------------------------------
%-------------------------------------Components of Method------------------------------------------------
%---------------------------------------------------------------------------------------------------------
\subsection{Discriminative Reconstruction Error Term}
To learn a representative and discriminative structured dictionary, each sub-dictionary $D_i$ should be able to well represent the dimensionality reduced samples from the $i$th class, but not other classes. To illustrate this idea mathematically, we rewrite $A_i$ as $A_i = [A_i^1 ; \dots; A_i^j ; \dots ; A_i^K]$, where $A_i^j$ is the representation coefficients of $P^T X_i$ over $D_j$. Our assumption implies that $A_i^i$ should have significant coefficients such that $\norm{P^T X_i - D_i A_i^i}_{F}^2$ is small, while for samples from class $j \, (j \neq i)$, $A_i^j$ should have nearly zero coefficients, such that $\norm{D_j A_i^j}_{F}^2$ is as small as possible. Moreover, the whole dictionary $D$ should well represent dimensionality reduced samples from any class, which implies the minimization of $\norm{P^T X_i - D A_i}_{F}^2$ in our model. 

\begin{figure}[t]
\centering
\subfloat[No incoherence penalty]{\includegraphics[width=3.5cm,keepaspectratio]{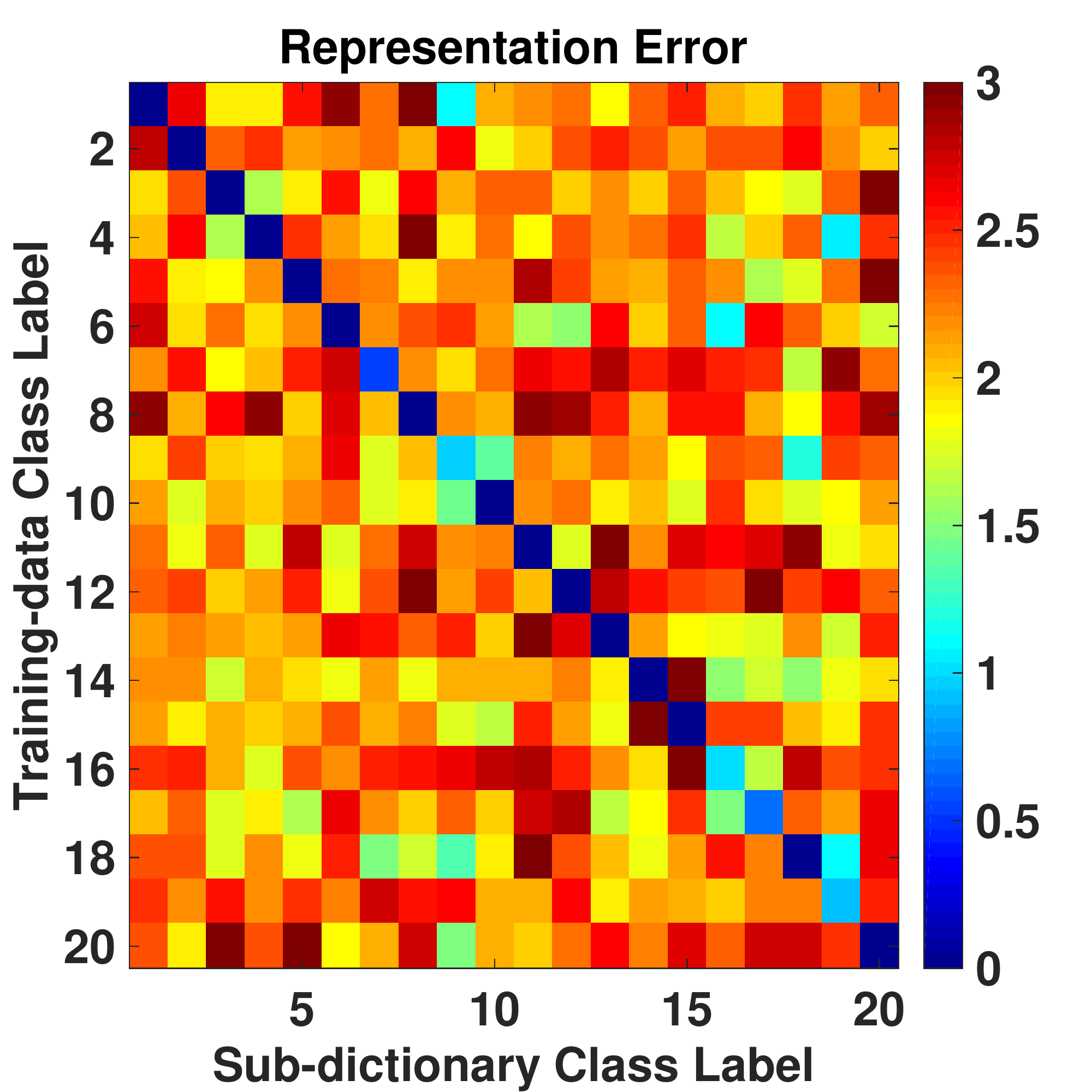}\label{fig:Error3}}
\hspace{6pt} 
\subfloat[With incoherence penalty]{\includegraphics[width=3.5cm,keepaspectratio]{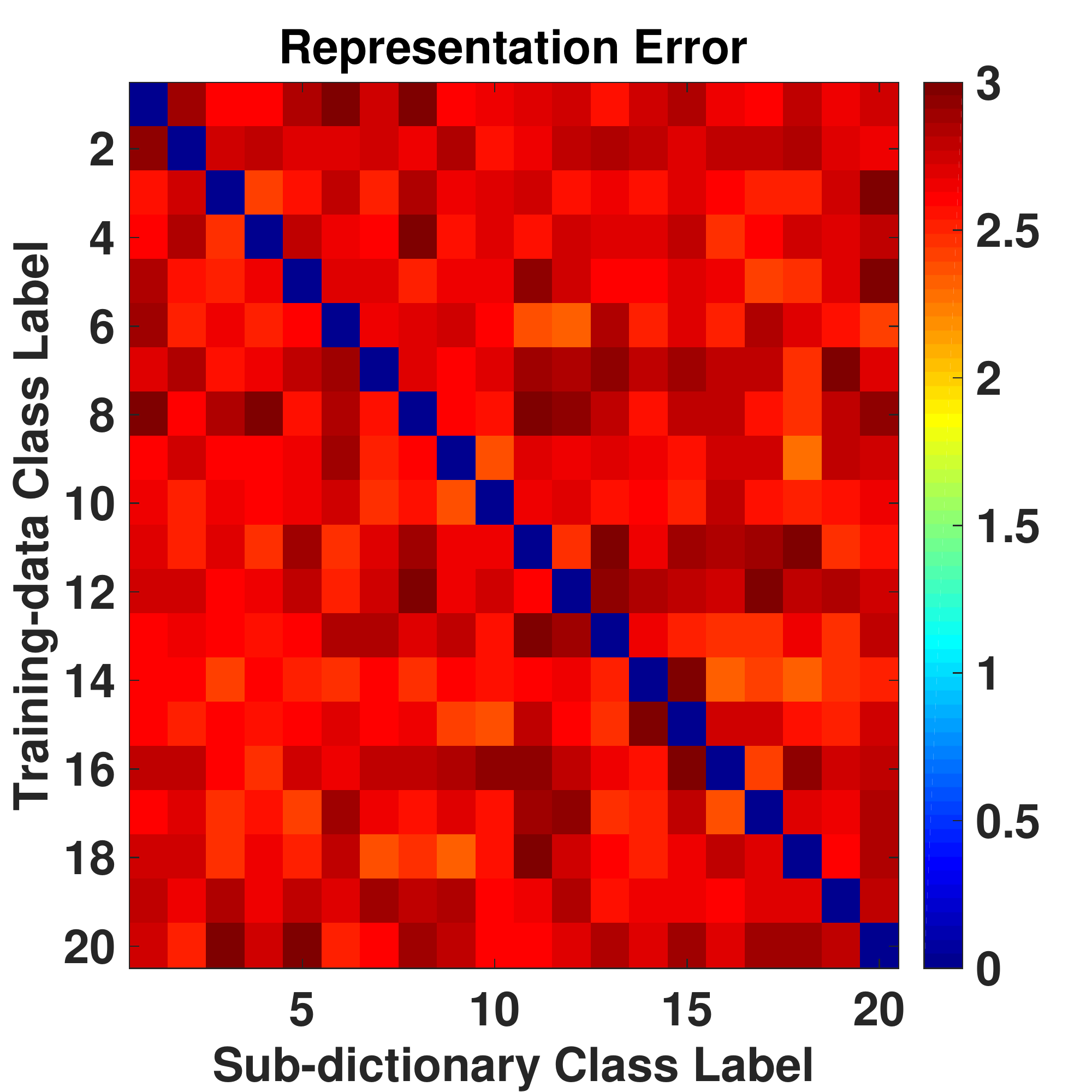}\label{fig:Error4}}
\caption{The role of the structural incoherence penalty term on a subset of the Caltech-101 dataset}
\label{fig:Rep_Error}
\vspace{-1.5em}
\end{figure}

Moreover, the common components of the samples in a dataset can be shared by a few or all the classes. Information redundancy in the original data leads to redundancy in the learned sub-dictionaries. So, in addition to the requirements of desirable discriminative reconstruction capability, we also need to promote incoherence among sub-dictionaries. We provide a structural incoherence constraint for sub-dictionaries as $\norm{D_i^T D_j}_{F}^2$ for $i \neq j$. Thus, the discriminative reconstruction term is defined as:
\begin{gather}
R(P,D,A) = \sum\limits_{i=1}^K \Big( \norm[\big]{P^T X_i - D A_i}_{F}^2 + \norm[\big]{P^T X_i - D_i A_i^i}_{F}^2
\label{eq2} \\ \notag
+ \sum_{\substack{j=1, j \neq i}}^K \norm[\big]{D_j A_i^j}_{F}^2 + \sum_{\substack{j=1, j \neq i}}^K \norm[\big]{D_i^T D_j}_{F}^2 \Big)
\end{gather}
Here, we use a subset of the Caltech-101 object dataset~\cite{Caltech} to better illustrate the role of incoherence penalty term. This dataset is known for imaging variations such as scale, viewpoint, lighting and background. The subset includes $20$ first classes with $20$ training samples per class. We learn the dictionary by using the first three terms and all terms of $R(P,D,A)$ and show the representation residuals of the training data over each sub-dictionary in Figures~\ref{fig:Error3} and~\ref{fig:Error4}, respectively. One can see that by using only the first three term in Equation~\eqref{eq2}, some training data may have big representation residuals over their associated sub-dictionaries because they can be partially represented by other sub-dictionaries. By adding incoherence terms in Equation~\eqref{eq2}, $D_i$ will have the minimal representation residual for $X_i$ and the redundancy among sub-dictionaries would be reduced effectively.
%&&&&&&&&&&&&&&&&&&&&&&&&&&&&&&&&&&&&&&&&&&&&&&&&&&&&&&&&&&&&&&&&&&&&&&&&&&&&&&&&&&&&&&&&&&&&&&&&&&&&&&&&
%---------------------------------------- Representation Error ------------------------------------------
%--------------------------------------------------------------------------------------------------------
%&&&&&&&&&&&&&&&&&&&&&&&&&&&&&&&&&&&&&&&&&&&&&&&&&&&&&&&&&&&&&&&&&&&&&&&&&&&&&&&&&&&&&&&&&&&&&&&&&&&&&&&&
\subsection{Graph-based Coding Coefficient Term}
To further increase the discrimination capability of dictionary $D$, we enforce the coding coefficient matrix $A$ to be discriminative. Intuitively, the discrimination can be assessed by the similarity of pairs of coding vectors from the same class and the dissimilarity of pairs of coding vectors from the different classes. This can be achieved by constructing a \textit{coefficient} graph and maximizing the intra-class compactness and inter-class separability of coding coefficients through proper definition of graph weights. We rewrite the sparse representation as $A=\{ \alpha_1, \alpha_2, \dots , \alpha_N \}$ where $\alpha_i$ is the coefficient vector of $x_i$ and $N$ is the number of training samples. 

First, using LR matrix recovery, the matrix of the data samples in the $i$th class, $X_i$, is decomposed into a LR matrix $L_i$ and sparse noise $E_i$ by the following optimization problem:
\begin{gather}
\min_{L_i,E_i} \> \norm{L_i}_{*} + \eta \norm{E_i}_{1} \quad s.t. \quad X_i=L_i+E_i \quad \forall i=1 \dots K
\label{eq3}
\end{gather}
Then, the weight matrix of the \textit{coefficient} graph, $W^c$, is defined as follows:
\begin{gather}
W^c_{ij} = 
\begin{dcases}
   1 & \, \parbox[c]{.8\columnwidth} {if $L(x_i) \in \mathcal N_{k1}(L(x_j))$ or $L(x_j) \in \mathcal N_{k1}(L(x_i))$ \hspace*{0.5cm} and \, $\textit{l}(x_i) = \textit{l}(x_j)$}\\
    \\
  -1  & \, \parbox[c]{.8\columnwidth} {if $L(x_i) \in \mathcal N_{k2}(L(x_j))$ or $L(x_j) \in \mathcal N_{k2}(L(x_i))$ \hspace*{0.5cm} and \, $\textit{l}(x_i) \neq \textit{l}(x_j)$}\\
    \\
    0       & \quad \text{otherwise}\\
\end{dcases}
\label{eq4}
\end{gather}
where $L(x_i)$ is the corresponding LR representation of image $x_i$ found by Equation~\eqref{eq3}, $\mathcal N_{k}(L(x_i))$ denotes the k-nearest neighbors of this representation and $\textit{l}(x_i)$ is the label of image $x_i$. Utilizing LR representation of images to determine their nearest neighbors, enables us to preserve this structure in the coefficient space, even if the images are occluded or corrupted. It is reasonable to use the weighted sum of the squared distances of pairs of coding vectors as an indicator of discrimination capability, resulting in the discriminative coefficient term as:
\begin{gather}
G(A) = \sum\limits_{i=1}^N \sum\limits_{j=1}^N \frac{1}{2} \norm[\big]{\alpha_i - \alpha_j}_{2}^2 \, W^c_{ij}
\label{eq5}
\end{gather}
This term ensures that the difference of the sparse codes of two images is minimized if they are from the same class and look similar, and the difference of the sparse codes of two images is maximized if they are from different classes and also look similar. Equation~\eqref{eq5} can be further simplified as:
\begin{gather}
G(A) = tr(A^T D^c A) - tr(A^T W^c A) = tr(A^T L^c A) 
\label{eq6}
\end{gather}
where $D^c$ is a diagonal matrix of column sums of $W^c$ as $D^c_{ii} = \sum\nolimits_{j} W^c_{ij}$ and $L^c$ is the Laplacian matrix as $L^c = D^c-W^c$. Interestingly,~\cite{SVGDL} showed that Fisher discrimination criterion, which is the most common discriminative coding coefficients term and originally adopted in~\cite{FDDL}, can be reformulated as a special case of the discrimination term in Equation~\eqref{eq5}.
%------------------------------------------------------------------------------------------------------------
\subsection{Low-rank Regularization}
The training samples in each class are linearly correlated in many situations and reside in a low-dimensional subspace. So, the sub-dictionary $D_i$, which is representing data from the $i$th class, is reasonably LR. Imposing LR constraints on sub-dictionaries would make them compact and mitigate the influence of noise and variations~\cite{D2L2R2}. To find the most compact bases, we need to minimize $\norm{D_i}_{*}$ for all classes in our optimization.
%------------------------------------------------------------------------------------------------------------
\subsection{Graph-based Projection Term}
We aim to learn a projection matrix that can preserve useful information and map the training samples to a discriminative space, where different classes are more discriminant toward each other, compared to the original space. Using the training data matrix $X$ and its corresponding class label set, the \textit{projection} graph is built.

First, we use Equation~\eqref{eq3} to find the LR representation of each image $x_i \in X$. Then, the weight matrix of the \textit{projection} graph, $W^p$, is defined as follows:
\begin{gather}
W^p_{ij} = 
\begin{dcases}
    d_1 & \, \parbox[c]{.8\columnwidth} {if $L(x_i) \in \mathcal N_{k1}(L(x_j))$ or $L(x_j) \in \mathcal N_{k1}(L(x_i))$ \hspace*{0.5cm} and \, $\textit{l}(x_i) = \textit{l}(x_j)$}\\
    \\
    d_2 & \, \parbox[c]{.8\columnwidth} {if $L(x_i) \in \mathcal N_{k2}(L(x_j))$ or $L(x_j) \in \mathcal N_{k2}(L(x_i))$ \hspace*{0.5cm} and \, $\textit{l}(x_i) \neq \textit{l}(x_j)$}\\
    \\
    0       & \quad \text{otherwise}\\
\end{dcases}
\label{eq7}
\end{gather}
where $L(x_i)$ is the corresponding LR representation of image $x_i$ found by Equation~\eqref{eq3}, $\mathcal N_{k}(L(x_i))$ denotes the k-nearest neighbors of this representation and $\textit{l}(x_i)$ is the label of $x_i$. 

To preserve the local geometrical structure in the projected space, one may naturally hope that, if two data points $x_i$ and $x_j$ are close in the intrinsic manifold, their corresponding low-dimensional embeddings $y_i$ and $y_j$ should also be close to each other. Ideally, similar data pairs which belong to different classes, in the original space should be far apart in the embedded space. As the first advantage of the \textit{projection} graph, it would enable us to preserve desirable relationship among training samples and penalize unfavorable relationship among them at the same time. This can be achieved by defining the weights $d_1(x_i,x_j) = exp (-\norm{L(x_i) - L(x_j)}^2/2t^2)$ and $d_2(x_i,x_j) = - exp (-\norm{L(x_i) - L(x_j)}^2/2t^2)$; where $t$ is considered as $1$ here. More importantly, these relationships should be persevered or penalized even if the images are heavily corrupted or occluded. Accordingly, we exploit the LR representation of images to determine their nearest neighbors and also to assign the weights of the matrix, rather than their original representation.

\begin{figure}[t]
\centering
\subfloat[Both LR and penalty]{\includegraphics[trim={0 0 0 1.1cm},clip,width=3.5cm,keepaspectratio]{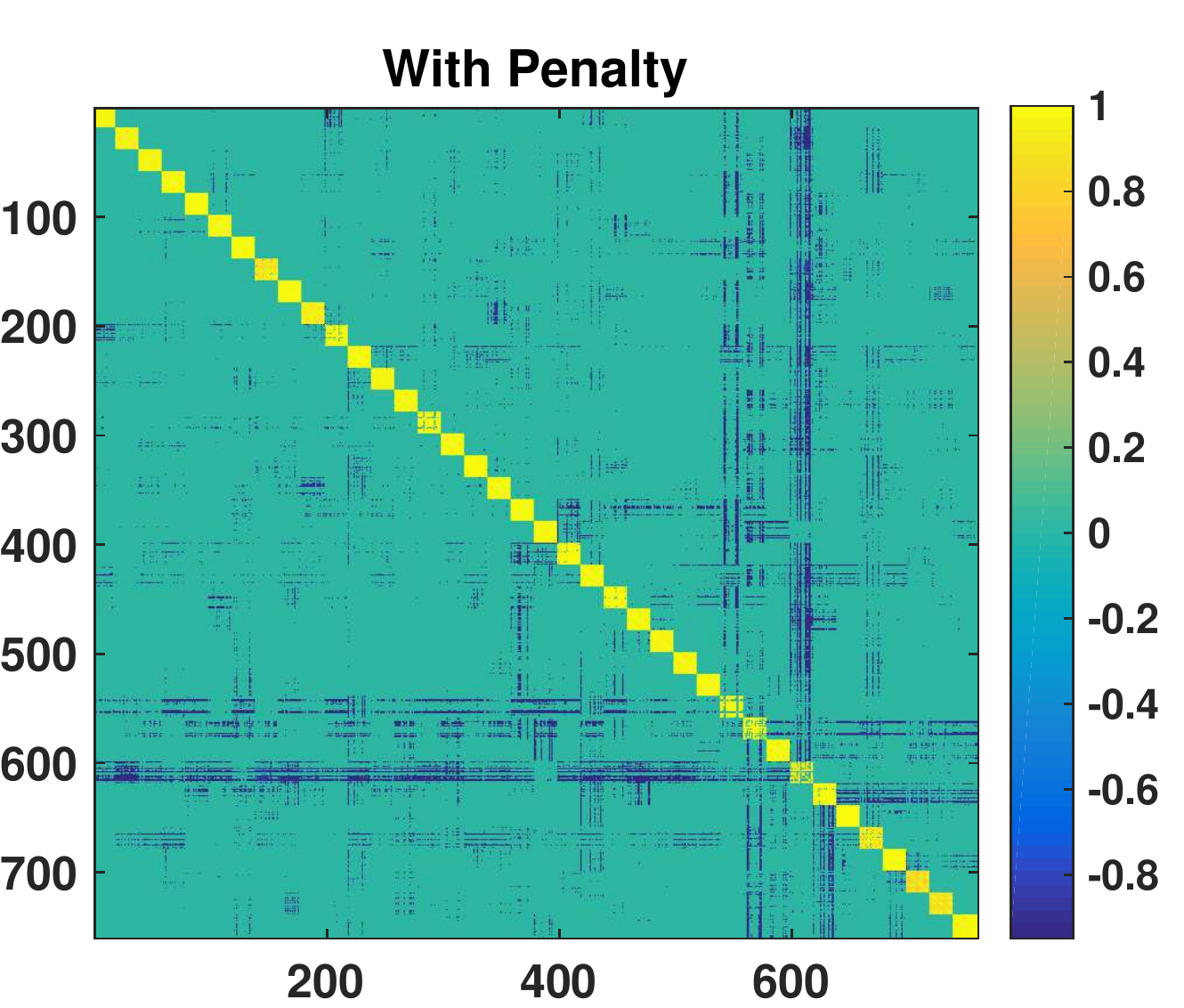}\label{fig:Penalty_LR}}
\hspace{4pt} 
\subfloat[Just LR, not penalty]{\includegraphics[trim={0 0 0 1.1cm},clip,width=3.5cm,keepaspectratio]{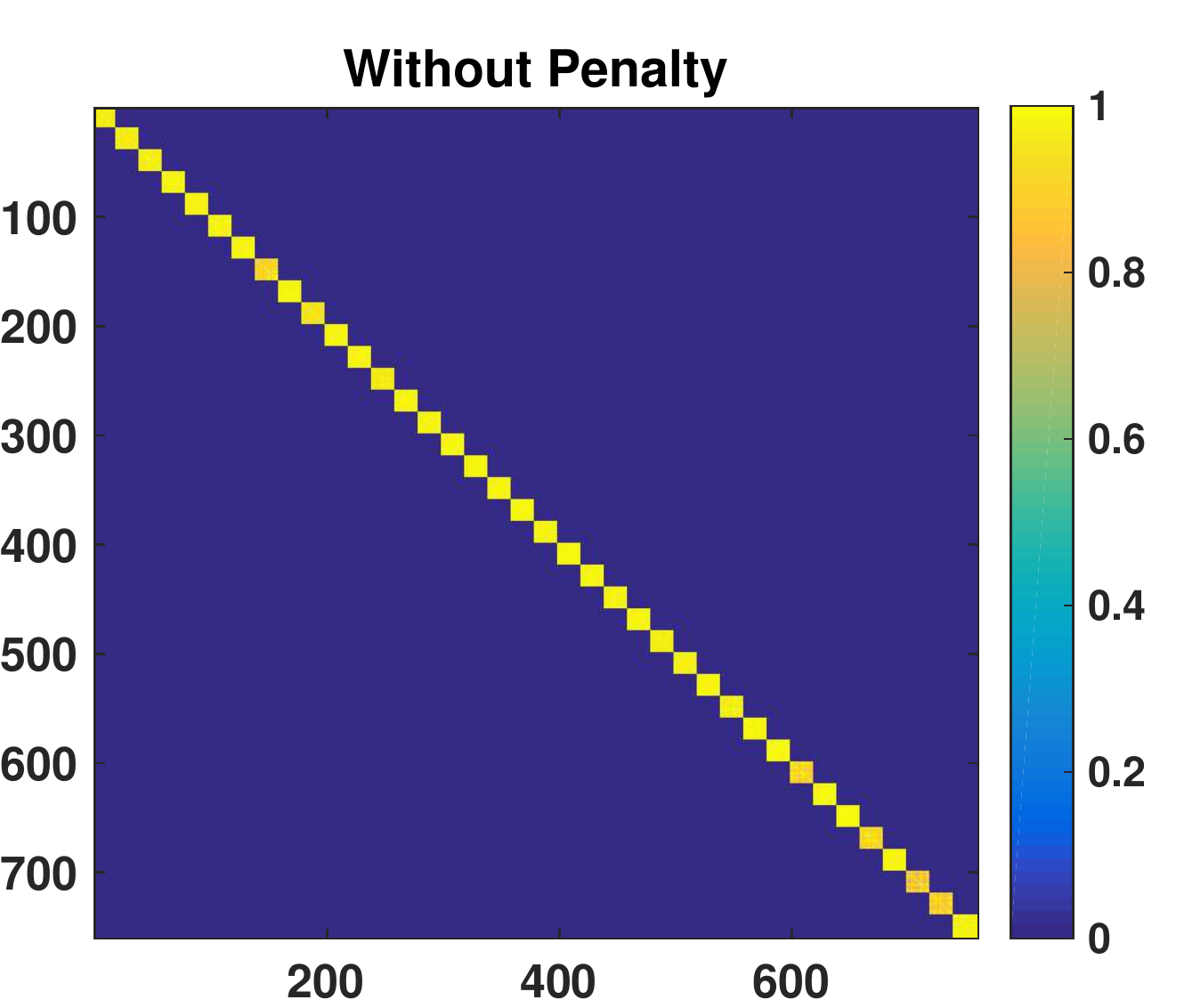}\label{fig:NoPenalty_LR}}
\vspace{2pt} 
\subfloat[Just penalty, not LR]{\includegraphics[trim={0 0 0 1.1cm},clip,width=3.5cm,keepaspectratio]{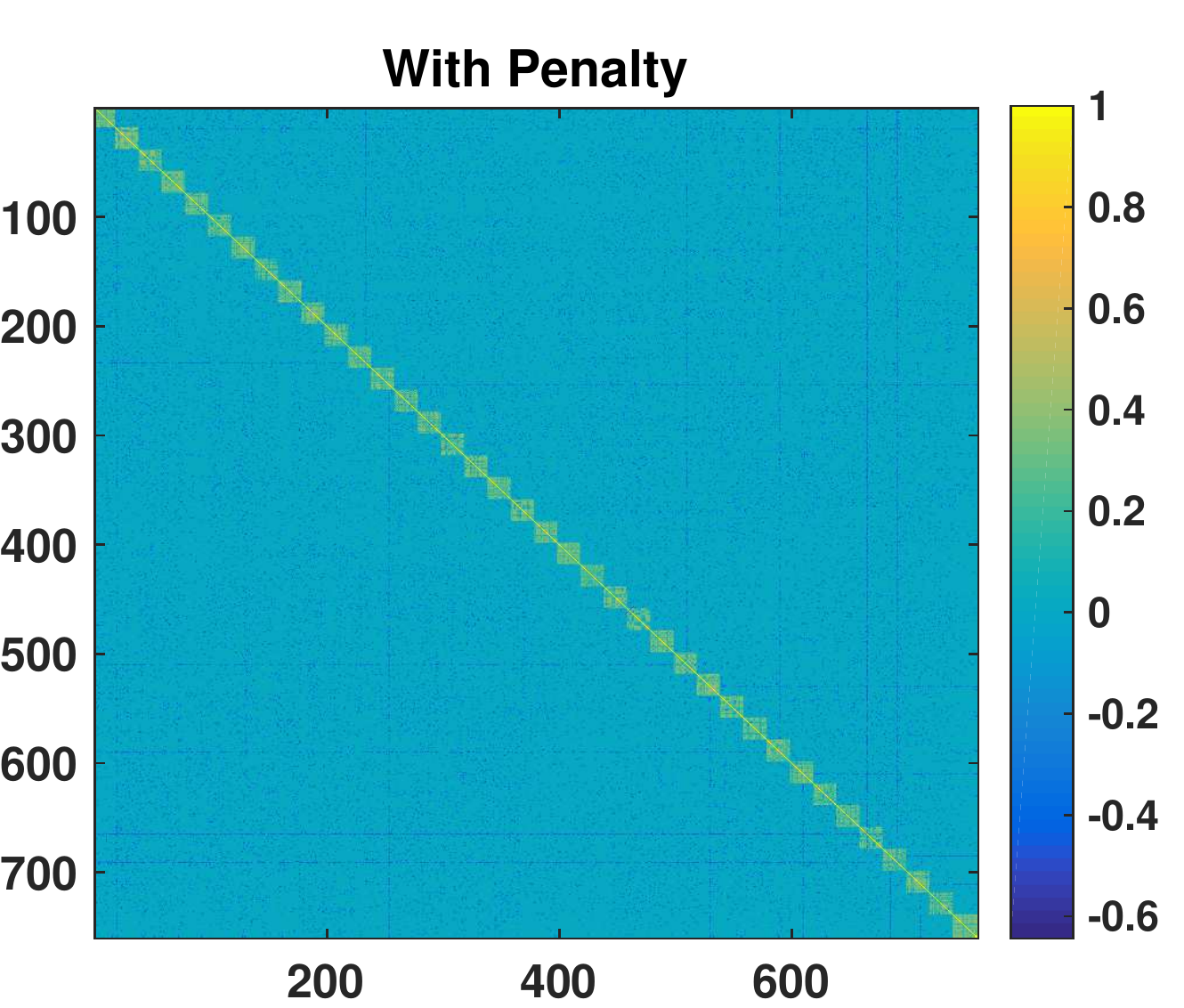}\label{fig:Penalty_NoLR}}
\hspace{4pt} 
\subfloat[Neither LR nor penalty]{\includegraphics[trim={0 0 0 1.1cm},clip,width=3.5cm,keepaspectratio]{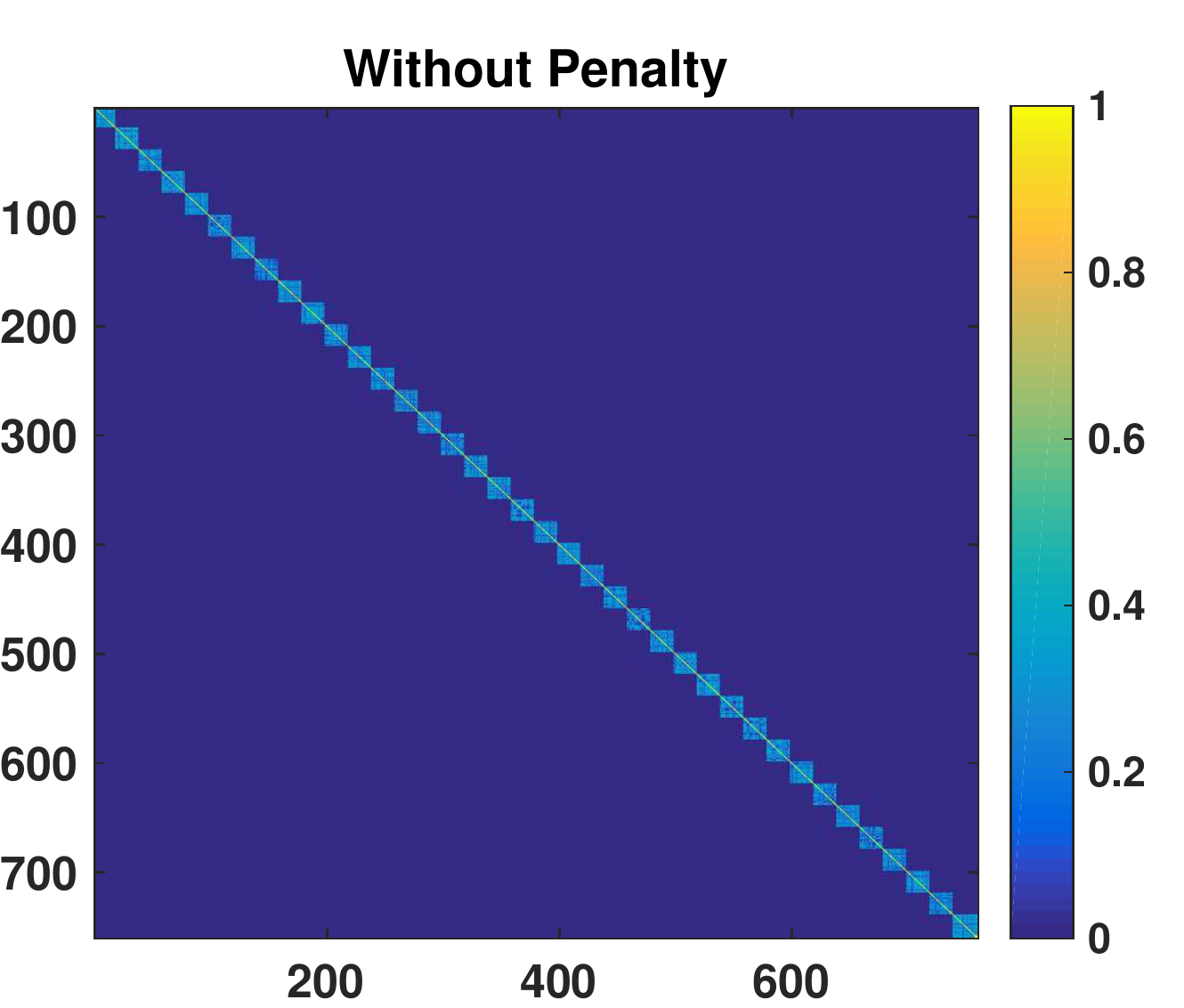}\label{fig:NoPenalty_NoLR}}
\caption{Comparison of weight matrices for 60\% pixel-corrupted images on the Extended YaleB dataset}
\vspace{-1.5em}
\end{figure}

Here, we illustrate the weight matrix $W^p$ for the Extended YaleB~\cite{Yale} face dataset, which is known for different illumination conditions and for extra challenge, we also simulate corruption by replacing $60\%$ of randomly selected pixels of each image with pixel value $255$. There are $38$ subjects in the dataset, and we randomly select $20$ training samples per class. Ideally the connecting weights between similar images from the same and different classes should be large and small, respectively. While the former is promoted, the latter should be penalized; hence, we can keep these relationships in the low-dimensional space. Figure~\ref{fig:Penalty_LR} shows the weight matrix found by Equation~\eqref{eq7}, which is confirming to the ideal case. There are two contributing factors in building this weight matrix, that we need to verify their importance. First, 
to spot the role of LR, we re-calculate the weight matrix, without utilizing LR representation (neither in neighborhood determination, nor in weights assignment) and demonstrate it in Figure~\ref{fig:Penalty_NoLR}. If we ignore the LR representation of images, corruption and illumination variations significantly deteriorates the weight matrix. Then, to verify the importance of penalizing unfavorable relations among similar training samples from different classes, we ignore the second condition of Equation~\eqref{eq7} and simply set all weights to zero; except those between similar pairs from the same class, which is obtained by $d_1(x_i,x_j)$. Figures~\ref{fig:NoPenalty_LR} and~\ref{fig:NoPenalty_NoLR} show these weight matrices with and without exploiting LR respectively. We observe that the our weight assignment in Figure~\ref{fig:Penalty_LR}, is much more discriminative and robust to variations, than the others.
%&&&&&&&&&&&&&&&&&&&&&&&&&&&&&&&&&&&&&&&&&&&&&&&&&&&&&&&&&&&&&&&&&&&&&&&&&&&&&&&&&&&&&&&&&&&&&&&&&&&&&&&&
%---------------------------------------- Weight Matrix -------------------------------------------------
%--------------------------------------------------------------------------------------------------------
%&&&&&&&&&&&&&&&&&&&&&&&&&&&&&&&&&&&&&&&&&&&&&&&&&&&&&&&&&&&&&&&&&&&&&&&&&&&&&&&&&&&&&&&&&&&&&&&&&&&&&&&&

We formulate the graph-based projection term as follows:
\begin{gather}
G(P) = \sum\limits_{i=1}^N \sum\limits_{j=1}^N \frac{1}{2} \norm[\big]{y_i - y_j}_{2}^2 \, W^p_{ij}
\label{eq8}
\end{gather}
Let $D^p$ be a diagonal matrix of column sums of $W^p$, $D^p_{ii} = \sum\nolimits_{j} W^p_{ij}$ and $L^p$ the Laplacian matrix as $L^p = D^p-W^p$. The cost function in Equation~\eqref{eq8} can be reduced to:
\begin{gather}
G(P) = tr(P^T X L^p X^T P) \quad s.t. \quad  P^T X D^p X^T P=I
\label{eq9}
\end{gather}
We note that the constraint $P^T X D^p X^T P =I$ removes the arbitrary scaling factor in the embedding. In order to make the constraint simpler, here we use the normalized graph Laplacian~\cite{Norm-Lap} as $\hat{L}^{p} = I - D^{p^-\frac {1}{2}} W^p D^{p^-\frac {1}{2}}$. Consequently, Equation~\eqref{eq9} is reformulated as:
\begin{gather}
G(P) = tr(P^T X \hat{L}^{p} X^T P)  \quad  s.t. \quad  P^T P=I
\label{eq10}
\end{gather}
By incorporating Equations~\eqref{eq2},~\eqref{eq6} and~\eqref{eq10} into the main optimization model, the JP-LRDL model is built.
%------------------------------------------------------------------------------------------------------------
\section{Optimization}
\label{sec:opt}
The objective function in Equation~\eqref{eq1} can be divided into three sub-problems to jointly learn dictionary $D$, projection $P$ and coding coefficients $A$. These sub-problems are optimized alternatively by updating one variable and fixing the other ones, through an iterative process. We summarize our proposed algorithm for JP-LRDL in Algorithm~\ref{alg2-whole} and discuss each sub-problem in details in the following subsections.
%------------------------------------------------------------------------------------------------------------
\subsection{Update of Coding Coefficients $A$}
Assuming that $D$ and $P$ are fixed, the objective function in Equation~\eqref{eq1} is further reduced to:
\begin{gather}
J_{(A)} = \argmin_{A}  \sum\limits_{i=1}^K \Big( \norm[\big]{P^T X_i - D A_i}_{F}^2 + \norm[\big]{P^T X_i - D_i A_i^i}_{F}^2
\label{eq11} \\ \notag
+ \sum_{\substack{j=1, j \neq i}}^K \norm[\big]{D_j A_i^j}_{F}^2 \Big) +  \lambda_{1} \, \norm[\big]{A}_{1} + \lambda_{2} \, tr(A^T L^c A) 
\end{gather}
We optimize $A_i$ class-by-class and meanwhile, make all other $A_j (j \neq i)$ fixed. As a result, Equation~\eqref{eq11} is simplified as:
\begin{gather}
J_{(A_i)} = \argmin_{A_i}  \norm[\big]{P^T X_i - D A_i}_{F}^2 + \norm[\big]{P^T X_i - D_i A_i^i}_{F}^2
\label{eq12} \\ \notag
+ \sum_{\substack{j=1, j \neq i}}^K \norm[\big]{D_j A_i^j}_{F}^2 +  \lambda_{1} \, \norm[\big]{A_i}_{1} + \lambda_{2} \, tr(A_i^T L^c A_i) 
\end{gather}
Following the work in~\cite{Feature-Sign}, we update $A_i$ one by one in the $i$th class. We define $\alpha_{i,p}$ as the coding coefficient of the $p$th sample in the $i$th class and optimize each $\alpha_{i,p}$ in $A_i$ alternatively, by fixing the encoding coefficients $\alpha_{j,p} (j \neq i)$ for other samples, and rewrite Equation~\eqref{eq12} as follows:
\begin{gather}
J_{(\alpha_{i,p})} = \argmin_{\alpha_{i,p}}  \norm[\big]{P^T X_i - D \alpha_{i,p}}_{F}^2 + \norm[\big]{P^T X_i - D_i \alpha_{i,p}^i}_{F}^2
\label{eq13} \\ \notag
+ \sum_{\substack{j=1, j \neq i}}^K \norm[\big]{D_j \alpha_{i,p}^j}_{F}^2 +  \lambda_{1} \, \norm[\big]{\alpha_{i,p}}_{1} + \lambda_{2} \, \mathcal Q(\alpha_{i,p})  
\end{gather}
where 
\begin{gather}
\mathcal Q(\alpha_{i,p}) = \lambda_{2} \Big(  \alpha_{i,p}^T A_i 	 L_p^c + (A_i  L_p^c )^T \alpha_{i,p} -  \alpha_{i,p}^T  L^c_{pp}  \alpha_{i,p} \Big)
\label{eq14}
\end{gather}
where $ L^c_p$ is the $p$th column of $L^c$, and $L^c_{pp}$ is the entry in the $p$th row and $p$th column of $L^c$. We then apply the feature-sign search algorithm~\cite{Feature-Sign} to solve $\alpha_{i,p}$.
%------------------------------------------------------------------------------------------------------------
%------------------------------------------------------------------------------------------------------------
\subsection{Update of Dictionary $D$}
Then, we optimize $D$ while $A$ and $P$ are fixed. We update $D_i$ class-by-class, by fixing all other $D_j ( j \neq i)$. When $D_i$ is updated, the coding coefficients of $ P^T X_i$ over $D_i$, \ie $A_i^i$ should also be updated to reflect this change. By ignoring irrelevant terms, the objective function of Equation~\eqref{eq1} then reduces to:
\begin{gather}
J_{(D_i,A_i^i)} = \argmin_{D_i,A_i^i} \Big\{ \norm[\big]{P^T X_i - D_i A_i^i - \sum\limits_{j=1 , \, j \neq i}^K D_j A_i^j}_{F}^2 
\label{eq15} \\ \notag
+ \norm[\big]{P^T X_i - D_i A_i^i}_{F}^2 + \sum\limits_{j=1 , \, j \neq i}^K \norm[\big]{D_j A_i^j}_{F}^2 
+ \sum_{\substack{j=1, j \neq i}}^K \norm[\big]{D_i^T D_j}_{F}^2 
\\ \notag
+ \lambda_3 \sum\limits_{i=1}^K  \norm[\big]{D_i}_{*} \Big\} 
\end{gather}
%------------------------------------------------------------------------------------------------------------
%---------------------------------Alg 1 ---------------------------------------------------------------------
%------------------------------------------------------------------------------------------------------------
\begin{algorithm}[h]
\footnotesize
%\scriptsize
\caption{Inexact ALM Algorithm for Equation~\eqref{eq19}}
\label{alg1-update-di}
\begin{algorithmic}[1]
\renewcommand{\algorithmicrequire}{\textbf{Input:}}
\renewcommand{\algorithmicensure}{\textbf{Output:}}
\algnewcommand{\Initialize}{\State \textbf{Initialize:}}
\Require Reduced-dimensionality data $P^TX_i$, Sub-dictionary $D_i$, parameters $\lambda_3, \beta, \lambda$
\Ensure $D_i, E_i, A_i^i$
\Initialize $\, J=0, \, E_i=0, \, T_1=0, \, T_2=0, \, T_3=0, \, \mu=10^{-6}, \, max_{\mu} = 10^{30}, \, \epsilon = 10^{-8}, \, \rho =1.1 $
\While {not converged}
    \State Fix other variables and update $Z$ as: 
    \vspace{2mm}
    \NoNumber{$ Z = \argmin_{Z} \, \Big( \frac{1}{\mu} \norm[\big]{Z}_{1} + \frac{1}{2} \norm[\big]{Z - (A_i^i + \frac{T3}{\mu})}_{F}^2 \Big) $}
    \vspace{2mm}    
    \State Fix other variables and update $A_i^i$ as: 
    \vspace{2mm}   
    \NoNumber{$A_i^i = \Big( D_i^TD_i +I \Big)^{-1} \Big( D_i^T(P^T X_i - E_i) +Z + \frac{D_i^T T_1 - T_3}{\mu} \Big)$}
    \vspace{2mm}    
    \State Fix other variables and update $J$ as:
    \vspace{2mm} 
    \NoNumber{$J = \argmin_{J} \, \Big( \frac{\lambda_3}{\mu} \norm[\big]{J}_{*} + \frac{1}{2} \norm[\big]{J - (D_i + \frac{T2}{\mu}}_{F}^2 \Big) $}
    \vspace{1mm}
    \NoNumber{Length normalization for each column in $J$}
    \vspace{2mm}
    \State Fix other variables and update $D_i$ as:
    \vspace{2mm}
    \NoNumber{ $\big( \frac{2 \lambda}{\mu} \sum\limits_{j=1 \atop j \neq i}^K D_j D_j^T \big) \, D_i + D_i \, \big(\frac{2 \lambda}{\mu} A_i^i {A_i^i}^T + A_i^i {A_i^i}^T + I \big)$} 
    \NoNumber{$ = \frac{2 \lambda}{\mu} \big( P^T X_i \> {A_i^i}^T - \sum\limits_{j=1 \atop j \neq i}^K D_j A_i^j {A_i^i}^T \big) + P^T X_i \> {A_i^i}^T - E_i {A_i^i}^T $}
    \NoNumber{$+ \, J + \frac{T_1 {A_i^i}^T - T_2}{\mu} $}
    \vspace{2mm}
    \NoNumber{The above problem is in the form of $HY + YQ = C$ and can be solved efficiently using algorithms in the literature~\cite{Eq-Di-1},~\cite{Eq-Di-2}.}
    \vspace{1mm}
    \NoNumber{Length normalization for each column in $D_i$}    
    \vspace{2mm}
    \State Fix other variables and update $E_i$ as:
    \vspace{2mm}   
    \NoNumber{$E_i = \argmin_{E_i} \, \Big( \frac{\beta}{\mu} \norm[\big]{E_i}_{2,1} + \frac{1}{2} \norm[\big]{E_i - (P^T X_i - D_i A_i^i + \frac{T1}{\mu})}_{F}^2 \Big) $}
    \vspace{2mm}   
    \State Update $T_1,T_2,T_3$ as: 
    \vspace{2mm}    
    \NoNumber {$\hspace{45pt} T_1 = T_1 + \mu (P^T X_i - D_i A_i^i - E_i)$}
    \NoNumber {$\hspace{45pt} T_2 = T_2 + \mu (D_i - J)$} 
    \NoNumber {$\hspace{45pt} T_3 = T_3 + \mu (A_i^i - Z)$} 
    \vspace{2mm}    
    \State Update $\mu$ as: $\mu = min(\rho \mu, max_{\mu})$    
    \vspace{2mm}    
    \State Check stopping conditions as: 
    \vspace{2mm}
    \NoNumber {$\hspace{-15pt} \norm[\big]{D_i - J}_{\infty} < \epsilon \quad \text{and} \quad \norm[\big]{P^T X_i - D_i A_i^i - E_i}_{\infty} < \epsilon \quad \text{and} \quad \norm[\big]{A_i^i - Z}_{\infty} < \epsilon $}
\EndWhile
\end{algorithmic}
\end{algorithm}
%------------------------------------------------------------------------------------------------------------
%------------------------------------------------------------------------------------------------------------
%------------------------------------------------------------------------------------------------------------
Denote
\begin{gather}
r(D_i) = \norm{P^T X_i - D_i A_i^i - \sum_{\substack{j=1, \, j \neq i}}^K D_j A_i^j}_{F}^2 \> +
\label{eq16} \\ \notag
\sum_{\substack{j=1, j \neq i}}^K \norm{D_j A_i^j}_{F}^2 + \sum_{\substack{j=1, j \neq i}}^K \norm{D_i^T D_j}_{F}^2
\end{gather}
Equation~\eqref{eq15}, can be converted to the following form:
\begin{gather}
\min_{D_i,A_i^i,E_i} \norm[\big]{A_i^i}_{1} + \lambda_3 \norm[\big]{D_i}_{*} + \beta \, \norm[\big]{E_i}_{2,1} + \lambda \, r(D_i)
\label{eq17} \\ \notag
\quad s.t. \,\,\, P^T X_i = D_i A_i^i + E_i
\end{gather}
Here, we first introduce the sparse error term $E$ and adopt $\norm{E_i}_{2,1}$ to characterize it, since we want to model the sample-specific corruption and outliers. This norm encourages the columns of $E$ to be zero, which is consistent with our assumption in the paper, that some vectors in data are corrupted. Moreover, we enforce sparsity on $\norm{A_i^i}_{1}$ to both avoid the trivial solution and keep sparsity constraint on coding coefficients. Then, to facilitate the optimization, we introduce two relaxation variables $J$ and $Z$ and then Equation~\eqref{eq17} can be rewritten as:
\begin{gather}
\min_{D_i,A_i^i,E_i} \norm[\big]{Z}_{1} + \lambda_3 \norm[\big]{J}_{*} + \beta \, \norm[\big]{E_i}_{2,1} + \lambda \, r(D_i)
\label{eq18} \\ \notag
s.t. \quad P^T X_i = D_i A_i^i + E_i , \> D_i = J, \> A_i^i = Z 
\end{gather}
The above problem can be solved by inexact Augmented Lagrange Multiplier (ALM) method~\cite{ALM-Proof}. The augmented Lagrangian function of Equation~\eqref{eq18} is:
\begin{gather}
\min_{D_i,A_i^i,E_i} \norm[\big]{Z}_{1} + \lambda_3 \norm[\big]{J}_{*} + \beta \, \norm[\big]{E_i}_{2,1} + \lambda \, r(D_i) 
\label{eq19} \\ \notag
+ tr \big[ T_1^T (P^T X_i - D_i A_i^i - E_i) \big] 
\\ \notag
+ tr \big[ T_2^T (D_i - J) \big] \hspace{50pt} 
\\ \notag
+ tr \big[ T_3^T (A_i^i - Z) \big] \hspace{50pt} 
\\ \notag
+ \frac{\mu}{2} \Big( \norm[\big]{P^T X_i - D_i A_i^i - E_i}_{F}^2 + \norm[\big]{D_i - J}_{F}^2 + \norm[\big]{A_i^i - Z}_{F}^2 \Big) 
\end{gather}
where $T_1$,$T_2$ and $T_3$ are Lagrange multipliers and $\mu$ is a balance parameter. The details of solving of Equation~\eqref{eq14} can be found in Algorithm~\ref{alg1-update-di}.
%------------------------------------------------------------------------------------------------------------
%------------------------------------------------------------------------------------------------------------
\subsection{Update of Projection Matrix $P$}
In order to solve for $P$, we keep $D$ and $A$ fixed. As a result, the objective function in Equation~\eqref{eq1} is then reduced to:
\begin{gather}
J_{(P)} = \argmin_{P} \Big\{  \sum\limits_{i=1}^K \big( \norm[\big]{P^T X_i - D A_i^i}_{F}^2 + \norm[\big]{P^T X_i - D_i A_i^i}_{F}^2 \big)  \label{eq20} \\ \notag 
+ \delta \, tr(P^T X \hat{L}^{p} X^T P) \Big\}
\quad s.t. \quad P^T P = I
\end{gather}
First, we rewrite the objective function in a more convenient form:
\begin{gather}
J_{(P)} = \argmin_{P} \Big\{ \norm[\big]{P^T X - \hat{D} \hat{Z}}_{F}^2 + \delta \, tr(P^T X \hat{L}^{p} X^T P) \Big\} 
\label{eq21} \\ \notag
s.t. \quad P^T P = I
\end{gather}
where $ \hat{D} = \Big[ [D,D_1] , [D,D_2], \dots, [D,D_K] \Big]$ and $\hat{Z}$ is a block-diagonal matrix, whose diagonal elements are formed as $\hat{Z}_{ii} = [A_i \, ; \, A_i^i] \> ; \> \forall i$. 

Because of the orthogonal constraint $P^T P = I$, we have $\norm{P^T X - \hat{D} \hat{Z}}_{F}^2 = tr(P^T \varphi(P) P) $, where $\varphi(P) = \big( X - P \hat{D} \hat{Z} \big) \big( X - P \hat{D} \hat{Z} \big)^T $. So, Equation~\eqref{eq16} is reformulated as:
\begin{gather}
J_{(P)} = \argmin_{P}  \> tr \Big( P^T \big( \, \varphi(P)+ \delta (X \hat{L}^{p} X^T) \big) P \Big) 
\label{eq22} \\ \notag
s.t. \quad P^T P = I
\end{gather}
To solve the above minimization, we iteratively update $P$ according to the projection matrix obtained in the previous iteration. Using singular value decomposition (SVD) technique, $[U,\Sigma,V^*] =  SVD \big( \varphi(P)+ \delta (X \hat{L}^{p} X^T) \big)$. Then, we can update $P$ as the $l$ eigenvectors in $U$ associated with the first $l$ largest eigenvalues of $\Sigma$, \ie, $P_t = U(1:l,:)$,  where $P_t$ is the projection matrix in the $t^{th}$ iteration. To avoid big changes in $P$ and make the optimization stable, we choose to update $P$ gradually in each iteration as following:
\begin{gather}
P_t =  P_{t-1} + \gamma \Big( U(1:l,:) - P_{t-1}) \Big) 
\label{eq23}
\end{gather}
$\gamma$ is a small positive constant to control the change of $P$ in consecutive iterations.
%&&&&&&&&&&&&&&&&&&&&&&&&&&&&&&&&&&&&&&&&&&&&&&&&&&&&&&&&&&&&&&&&&&&&&&&&&&&&&&&&&&&&&&&&&&&&&&&&&&&&&&&&
%---------------------------------Alg 2 -----------------------------------------------------------------
%--------------------------------------------------------------------------------------------------------
\begin{algorithm}[h]
\footnotesize
%\scriptsize
\caption{JP-LRDL Algorithm}
\label{alg2-whole}
\begin{algorithmic}[1]
\renewcommand{\algorithmicrequire}{\textbf{Input:}}
\algnewcommand{\Initialize}{\State \textbf{Initialize:}}
\algnewcommand{\Output}{\State \textbf{Output:}}
\Initialize 
\NoNumber{Projection $P$ as LPP~\cite{LPP} of $X$}
\NoNumber{Dictionary $D$; set the atoms of $D_i$ as the eigenvectors of $P^T X_i$}
    \vspace{1mm}    
    \State \textbf{Update the coefficient matrix $A$} 
    \vspace{1mm}
    \NoNumber{Fix $D,P$ and solve $A_i \> i=1,2, \dots, K$; one by one by solving Equation~\eqref{eq13} using Feature-sign search algorithm~\cite{Feature-Sign}.}
    \vspace{2mm}
    \State \textbf{Update the dictionary $D$} 
    \NoNumber{Fix $A,P$ and solve $D_i \> i=1,2, \dots, K$; one by one by solving Equation~\eqref{eq19} using inexact ALM algorithm~\ref{alg1-update-di}.}
    \vspace{2mm}
    \State \textbf{Update the projection $P$} 
    \NoNumber{Fix $D,A$ and solve $P$ by solving Equation~\eqref{eq23}.}
    \vspace{2mm}    
    \Output 
    \NoNumber{Return to step $2$ until the objective function values in consecutive iterations are close enough or the maximum number of iterations is reached. Then output $P$,$D$ and $A$.}
\end{algorithmic}
\end{algorithm}
%---------------------------------------------------------------------------------------------------------
%&&&&&&&&&&&&&&&&&&&&&&&&&&&&&&&&&&&&&&&&&&&&&&&&&&&&&&&&&&&&&&&&&&&&&&&&&&&&&&&&&&&&&&&&&&&&&&&&&&&&&&&&
%|||||||||||||||||||||||||||||||||||||||||||||||||||||||||||||||||||||||||||||||||||||||||||||||||||||||||
%||-----------------------------------------------------------------------------------------------------||
%||-----------------------------------------------------------------------------------------------------||
%||----------------------------------- Complexity/Convergence ------------------------------------------||
%||-----------------------------------------------------------------------------------------------------||
%||-----------------------------------------------------------------------------------------------------||
%|||||||||||||||||||||||||||||||||||||||||||||||||||||||||||||||||||||||||||||||||||||||||||||||||||||||||
\section{Complexity and Convergence Analysis}
\label{sec:comp}
\subsection{Time Complexity}
We analyze the time complexity of three sub-problems of JP-LRDL optimization as follows:
\begin{itemize}
\item Update coding coefficients: Feature-sign search algorithm~\cite{Feature-Sign} has a time complexity of $O(sC)$, where $s$ is the sparsity level of the optimal solution \ie the number of nonzero coefficients and $C$ is the dictionary size.
\item Update dictionary: The most time-consuming steps in Algorithm~\ref{alg1-update-di} are steps 1 and 3, which need SVD with cost $O(n^3)$, where $n$ is the number of training samples. Therefore, the total complexity of Algorithm~\ref{alg1-update-di} is $O(t_1 \, n^3)$, where $t_1$ is the number of iterations of this algorithm.
\item Update projection matrix: The most time-consuming step in updating $P$ is SVD; so, the time complexity of this step would be $O(n^3)$.
\end{itemize}
Hence, the total time complexity of JP-LRDL is $t_2 \, O(n^3+sC)$, where $t_2$ is the total number of iterations. 
%------------------------------------------------------------------------------------------------------------
\subsection{Convergence Analysis}
%&&&&&&&&&&&&&&&&&&&&&&&&&&&&&&&&&&&&&&&&&&&&&&&&&&&&&&&&&&&&&&&&&&&&&&&&&&&&&&&&&&&&&&&&&&&&&&&&&&&&&&&&
% --------------------------- Convergence ---------------------------------------------------------------
\begin{figure}[t]
\centering
\subfloat[]{\includegraphics[width=3.7cm,keepaspectratio]{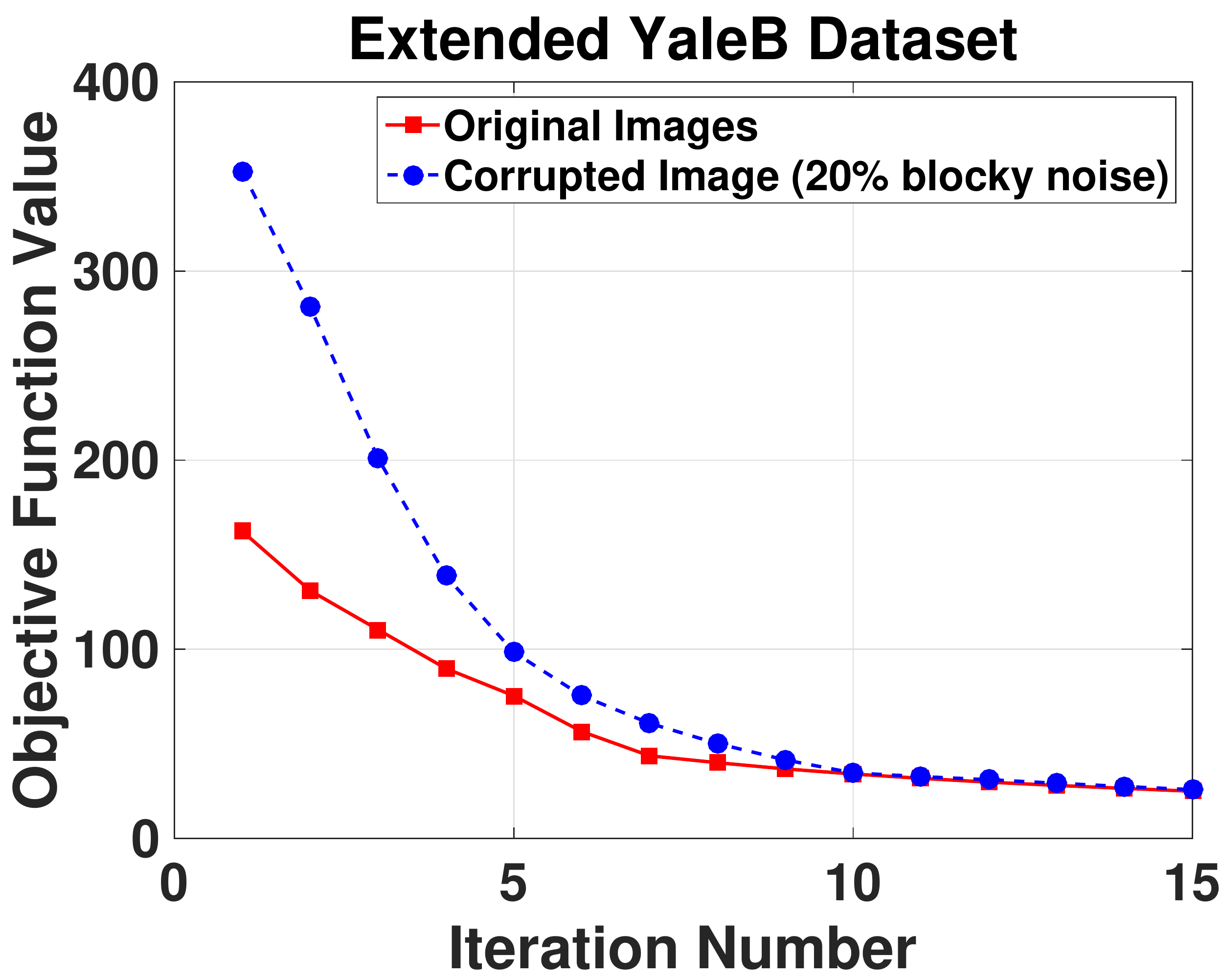} \label{fig:Yale_Convergence}}  
\hspace{2pt} 
\subfloat[]{\includegraphics[width=3.7cm,keepaspectratio]{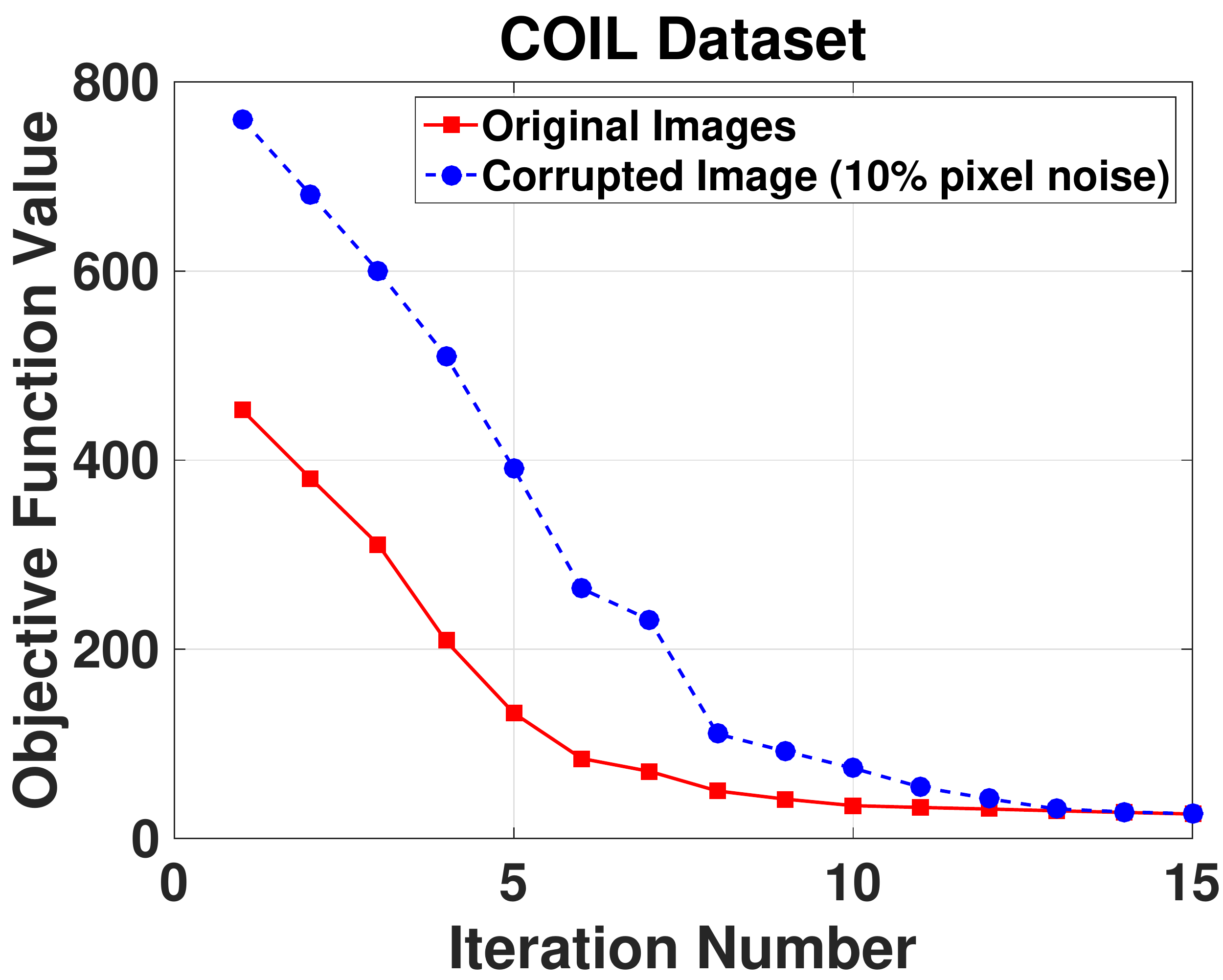} \label{fig:COIL_Convergence}}  
\vspace{-0.1em}
\subfloat[]{\includegraphics[width=6cm,keepaspectratio]{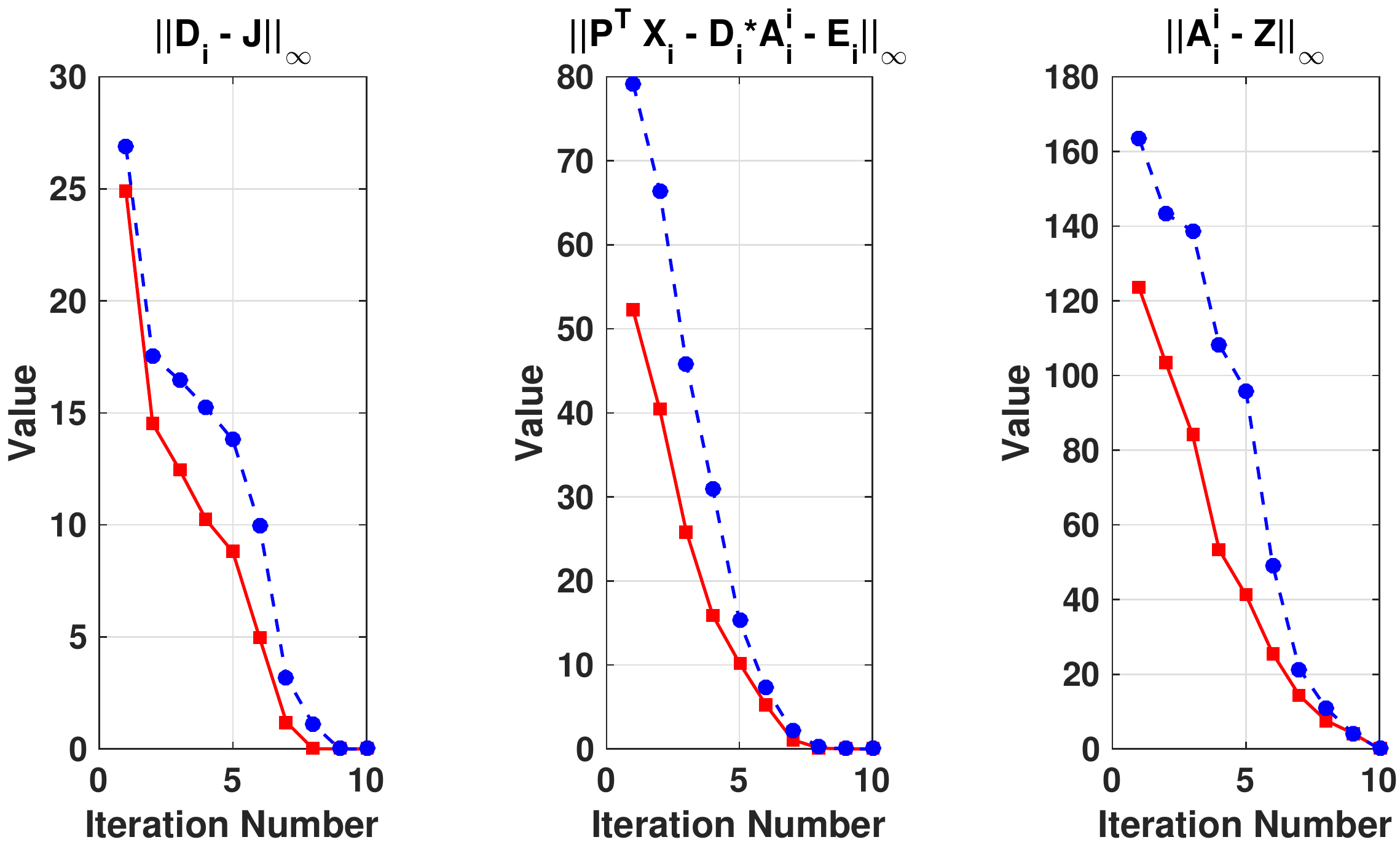}  \label{fig:Dic_Convergence}}  
\caption{The convergence of JP-LRDL on (a) the Extended YaleB dataset (b) the COIL dataset (c) The curve of stopping conditions of Algorithm~\ref{alg1-update-di} on the Extended YaleB dataset}
\vspace{-1.5em}
\end{figure} 
%--------------------------------------------------------------------------------------------------------
%&&&&&&&&&&&&&&&&&&&&&&&&&&&&&&&&&&&&&&&&&&&&&&&&&&&&&&&&&&&&&&&&&&&&&&&&&&&&&&&&&&&&&&&&&&&&&&&&&&&&&&&&&&&&
Although Equation~\eqref{eq1} is non-convex, the convergence of each sub-problem is guaranteed. For updating coding coefficients, we exploit feature-sign search algorithm, which~\cite{Feature-Sign} proved this algorithm converges to a global optimum of the optimization problem in a finite number of steps. For updating sub-dictionaries, we use inexact ALM as demonstrated in Algorithm~\ref{alg1-update-di}. The convergence of inexact ALM, with at most two blocks has been well studied and a proof to demonstrate its convergence property can be found in~\cite{ALM-Proof}. Liu \etal\cite{IALM-3block} also showed that there actually exist some guarantees for ensuring the convergence of inexact ALM with three or more blocks (here $Z$, $J$ and $E$). So, it could be well expected that Algorithm~\ref{alg1-update-di} has good convergence properties. Moreover, inexact ALM is known to generally perform well in reality, as illustrated in~\cite{ADM-Tutorial}. The convergence of updating the projection matrix in Equation~\eqref{eq23} has also been discussed in~\cite{JDDRDL}. 

In addition, Figures~\ref{fig:Yale_Convergence} and~\ref{fig:COIL_Convergence} illustrate the convergence curves of JP-LRDL on the original and corrupted images of the Extended YaleB face~\cite{Yale} and the COIL object~\cite{COIL} datasets, respectively. It can be observed that JP-LRDL converges efficiently and after several iterations, the values of objective function becomes stable, such that local solutions cannot make the problem unpredictable. Although the objective function value on corrupted images is larger than that of original ones, the function converges very well after some iterations in both cases. Additionally, Figure~\ref{fig:Dic_Convergence} demonstrates the value of $\norm{D_i - J}_{\infty}$, $\norm{P^T X_i - D_i A_i^i - E_i}_{\infty}$ and $\norm{A_i^i - Z}_{\infty}$, which are the stopping conditions of Algorithm~\ref{alg1-update-di}, on the original and corrupted images of the Extended YaleB dataset. We observe that inexact ALM efficiently convergences through few iterations in both cases.
%|||||||||||||||||||||||||||||||||||||||||||||||||||||||||||||||||||||||||||||||||||||||||||||||||||||||||
%||-----------------------------------------------------------------------------------------------------||
%||-----------------------------------------------------------------------------------------------------||
%||----------------------------------- Classification --------------------------------------------------||
%||-----------------------------------------------------------------------------------------------------||
%||-----------------------------------------------------------------------------------------------------||
%|||||||||||||||||||||||||||||||||||||||||||||||||||||||||||||||||||||||||||||||||||||||||||||||||||||||||
\section{The Classification Scheme}
\label{sec:classification}
Once $D$ and $P$ are learned, they could be used to represent a query sample $x_{test}$ and find its corresponding label. The test sample is projected into the low-dimensional space and coded over $D$ by solving the following equation:
\begin{gather}
\hat{a} = \argmin_{a} \big\{ \norm[\big]{P^T x_{test} - D a}_{2}^2 + \xi \norm[\big]{a}_{1} \big\}
\label{eq24}
\end{gather}
$\xi$ is a positive scalar and the coding vector $\hat{a}$ can be written as $\hat{a} = [\hat{a}_1, \hat{a}_2, \dots \hat{a}_K]$ where $\hat{a}_i$ is the coefficient sub-vector associated with sub-dictionary $D_i$. The representation residual for the $i$th class is calculated as:
\begin{gather}
e_i = \norm[\big]{P^T x_{test} - D_i \hat{a}_i}_{2}^2 + \omega \norm[\big]{\hat{a} - m_i}_{2}^2 
\label{eq25}
\end{gather}
where $\omega$ is a preset weight. Finally, the identity of testing sample is determined by $identity(x_{test}) = \argmin_{i} \{e_i\}$.
%|||||||||||||||||||||||||||||||||||||||||||||||||||||||||||||||||||||||||||||||||||||||||||||||||||||||||
%||-----------------------------------------------------------------------------------------------------||
%||-----------------------------------------------------------------------------------------------------||
%||----------------------------------- Experiments -----------------------------------------------------||
%||-----------------------------------------------------------------------------------------------------||
%||-----------------------------------------------------------------------------------------------------||
%|||||||||||||||||||||||||||||||||||||||||||||||||||||||||||||||||||||||||||||||||||||||||||||||||||||||||
%&&&&&&&&&&&&&&&&&&&&&&&&&&&&&&&&&&&&&&&&&&&&&&&&&&&&&&&&&&&&&&&&&&&&&&&&&&&&&&&&&&&&&&&&&&&&&&&&&&&&&&&&
% --------------------------- Parameters Curve /Running Time --------------------------------------------
\begin{figure}[b]
\centering
\subfloat[]{\includegraphics[width=4.2cm,keepaspectratio]{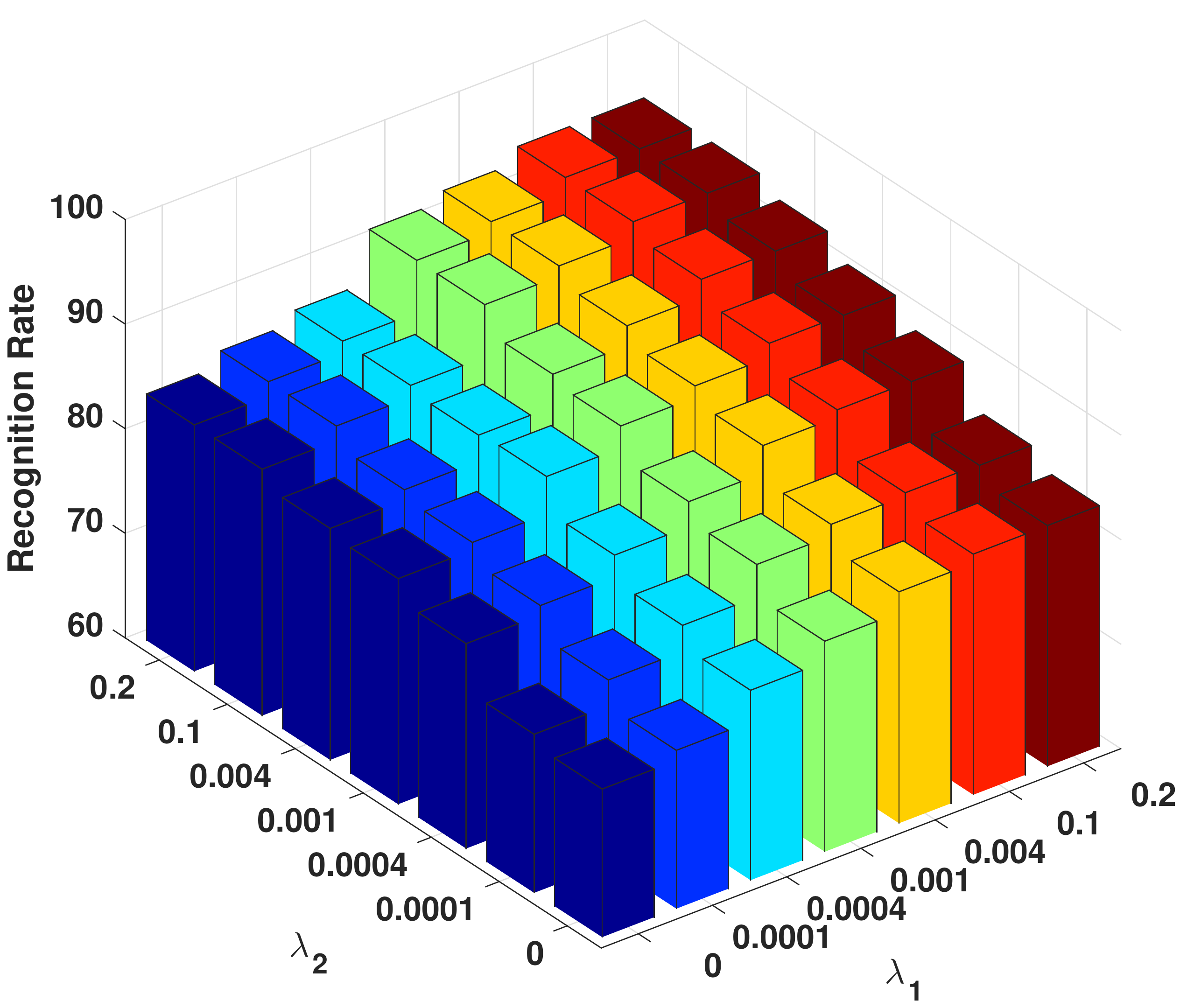} \label{fig:Lambda-Curves}}  
\hspace{2pt} 
\subfloat[]{\includegraphics[width=4.2cm,keepaspectratio]{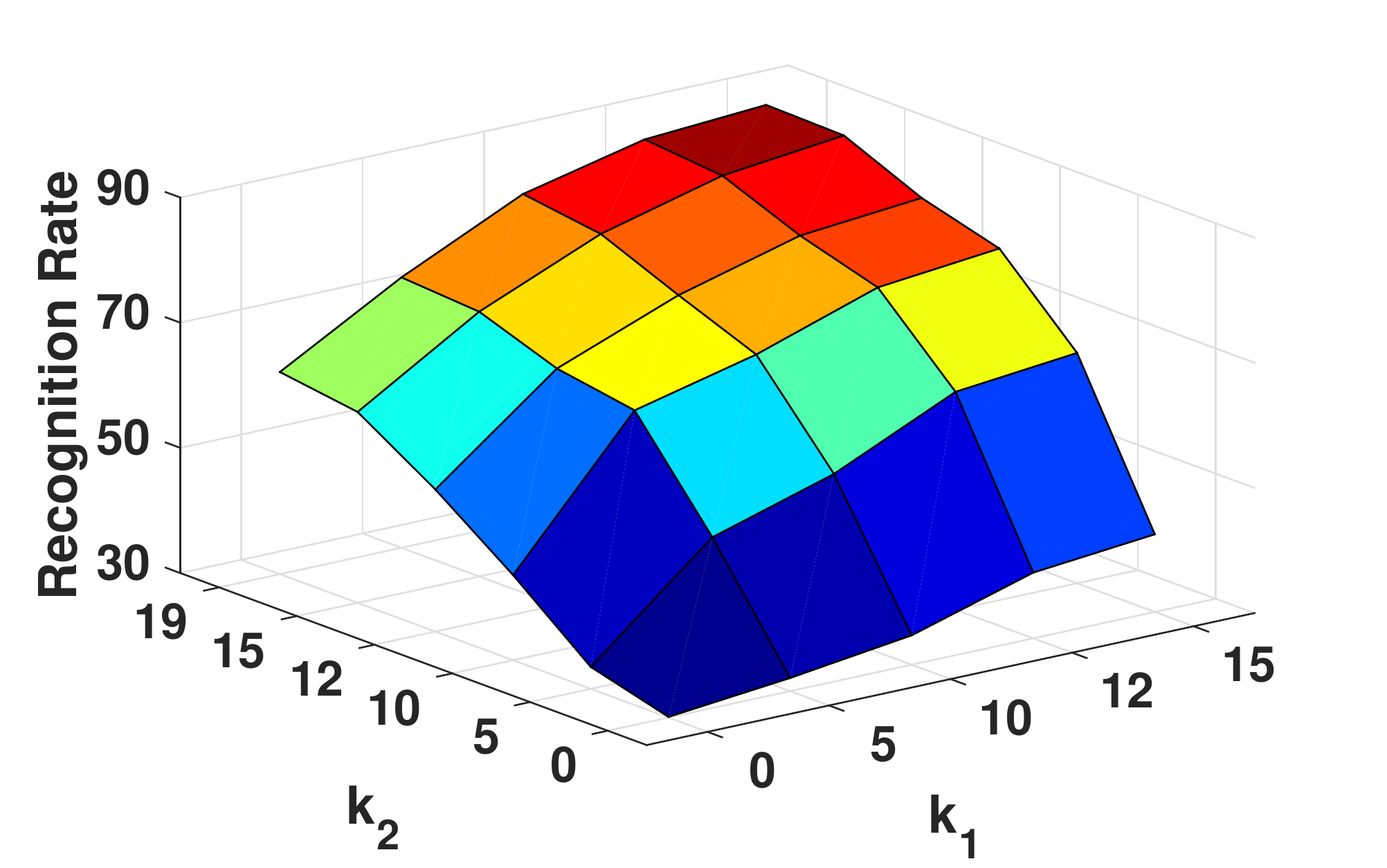} \label{fig:K-Curves}}  
\vspace{0.2em} 
\subfloat[]{\includegraphics[width=4.2cm,keepaspectratio]{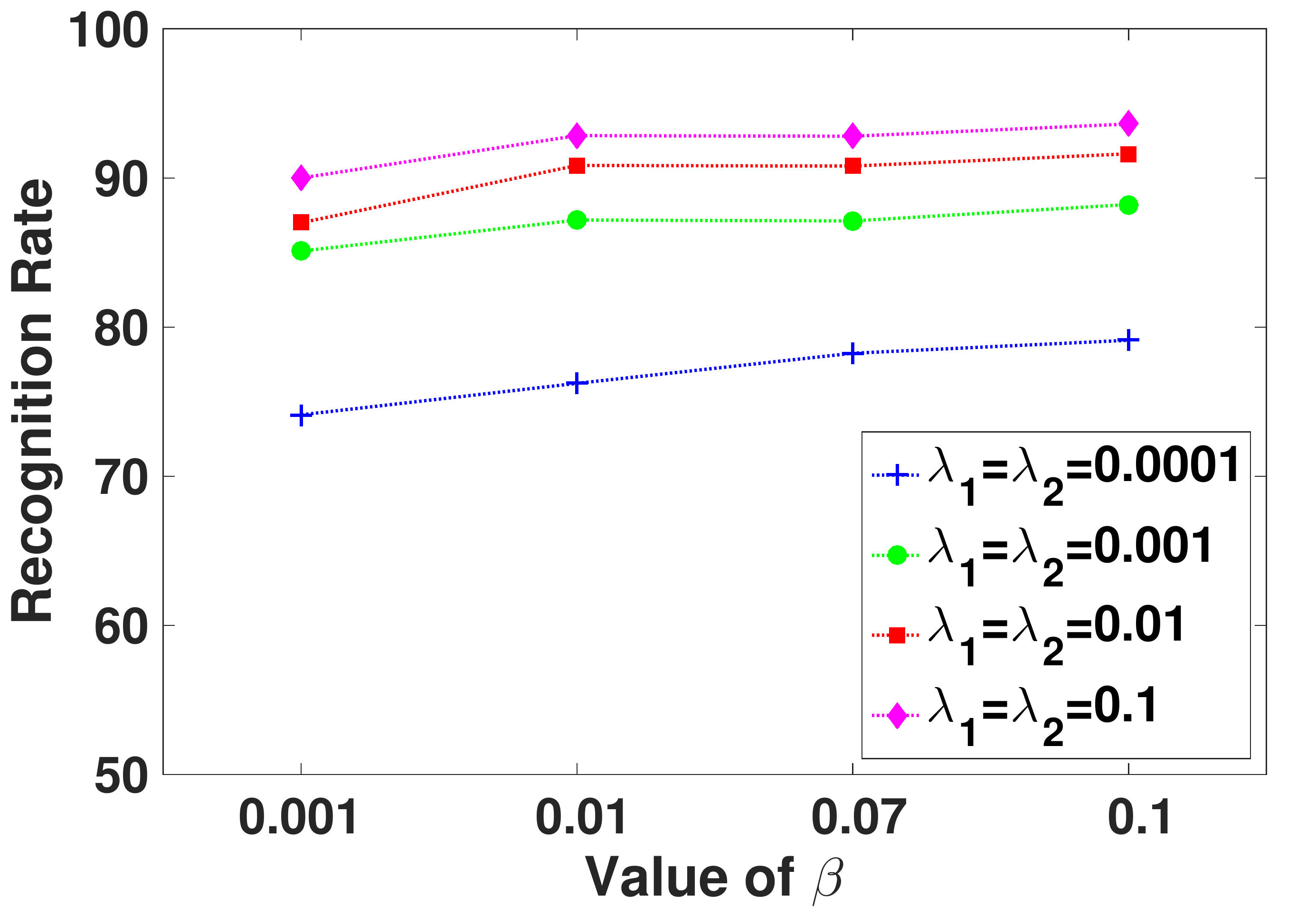} \label{fig:Beta-Curve}}  
\hspace{2pt} 
\subfloat[]{\includegraphics[width=4.2cm,keepaspectratio]{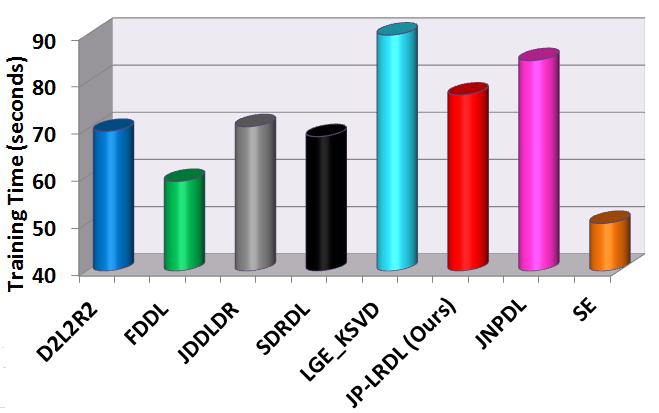} \label{fig:Train-Time}}  
\caption{Performance of JP-LRDL on different values of (a) $\lambda_1$, $\lambda_2$ parameters on the AR dataset (b) neighborhood size on the Extended YaleB dataset (c) $\beta$ parameter on the AR dataset and (d) Average training time on the Extended YaleB dataset}
\vspace{-1.5em}
\end{figure} 
%--------------------------------------------------------------------------------------------------------
%&&&&&&&&&&&&&&&&&&&&&&&&&&&&&&&&&&&&&&&&&&&&&&&&&&&&&&&&&&&&&&&&&&&&&&&&&&&&&&&&&&&&&&&&&&&&&&&&&&&&&&&&
\section{Experimental Results}
\label{sec:experiment}
The performance of JP-LRDL method is evaluated on various classification tasks. We compare our method with several related methods. FDDL~\cite{FDDL} and D\textsuperscript{2}L\textsuperscript{2}R\textsuperscript{2}~\cite{D2L2R2} are representative of conventional DL and low-rank DL methods, respectively. We also compare JP-LRDL with joint DR and DL methods including JNPDL~\cite{JNPDL}, SDRDL~\cite{Simul-DL}, SE~\cite{SE}, LGE-KSVD~\cite{LGE-KSVD} and JDDRDL~\cite{JDDRDL}. Since SE can obtain at most $K$ (number of classes) features in the reduced space, it would be excluded from the experiment which is not applicable.

We evaluate the performance of our approach and related methods on the different face and object datasets. For constructing the training set, we select images (or their corresponding features) randomly and the random selection process is repeated $10$ times and we report the average recognition rates for all methods. We set the number of dictionary atoms of each class as training size. Also, we set the maximum iteration of all the iterative methods as $10$. For all the competing methods, we use their original settings and all the hyper-parameters are found by 5-fold cross validation.
%--------------------------------------------------------------------------------------------------------
\subsection{Parameters Selection}
There are seven parameters in our model, which need to be tuned: $\lambda_1$, $\lambda_2$, $\lambda_3$, $\delta$ in Equation~\eqref{eq1}, $\beta$, $\lambda$ in Equation~\eqref{eq17} and $\gamma$ in Equation~\eqref{eq23}. The tuning parameters of JP-LRDL are chosen by 5-fold cross validation. However, we found out that changing $\lambda_3$, $\delta$ and $\lambda$ would not affect the results that much and we set them as $1$. There are also two parameters in classification phase as $\xi, \omega$, that we search for their best values in a small set $\{0.001, 0.01,0.1\}$. 

First, to investigate how sensitive the $\lambda_1$, $\lambda_2$ parameters are, we set the value of $\beta=0.1, \gamma=0.1$ and then explore the effects of the other two parameters. Figure~\ref{fig:Lambda-Curves} shows the recognition rate versus different values of these two parameters by fixing $\beta,\gamma$ as $0.1$ on the AR face dataset (Mixed scenario as explained in Section~\ref{sec:experiment}). For each pair of parameters, we average the results over $10$ random runs. We observe the accuracy reaches a plateau as either $\lambda_1$ or $\lambda_2$ grow from $0.1$. This trend is mostly similar in all evaluated datasets. We notice that when $\lambda_1=0$, the accuracy drops remarkably, which shows the importance of the sparsity of the coefficients. 
Figure~\ref{fig:Beta-Curve} also illustrates the recognition rate versus the value of $\beta$, under four different pair values of $\lambda_1$, $\lambda_2$ on the AR dataset. We note that JP-LRDL performs well in a reasonable range of $\beta$ parameter and the highest accuracy belongs to the $\lambda_1 =\lambda_2=0.1$, which is consistent with the results from Figure~\ref{fig:Lambda-Curves}. We set the $\gamma$ parameter as $0.1$ in the experiments.

For both \textit{coefficient} and \textit{projection} graphs, we set the neoghborhood size for similar and different classes as $k_1 = min\{n_{i}-1,15\}$ and $k_2=n_i-1$, where $n_i$ is the number of training samples in class $i$. Figure~\ref{fig:K-Curves} shows the classification results varying the neighborhood size $k_1$, $k_2$ on the Extended YaleB dataset. In this experiment, images are corrupted by $30\%$ occlusion and we randomly select $n_i = 20$ images per class. As the number of neighbors increases, JP-LRDL achieves better results and using relatively few neighborhoods, remarkably degrades the classification accuracy.
%-------------------------------------------------------------------------------------------------------
%&&&&&&&&&&&&&&&&&&&&&&&&&&&&&&&&&&&&&&&&&&&&&&&&&&&&&&&&&&&&&&&&&&&&&&&&&&&&&&&&&&&&&&&&&&&&&&&&&&&&&&&&
% --------------------------- YaleB recognition results -------------------------------------------------
\begin{figure}[t]
\centering
\subfloat[]{\includegraphics[width=4.2cm,keepaspectratio]{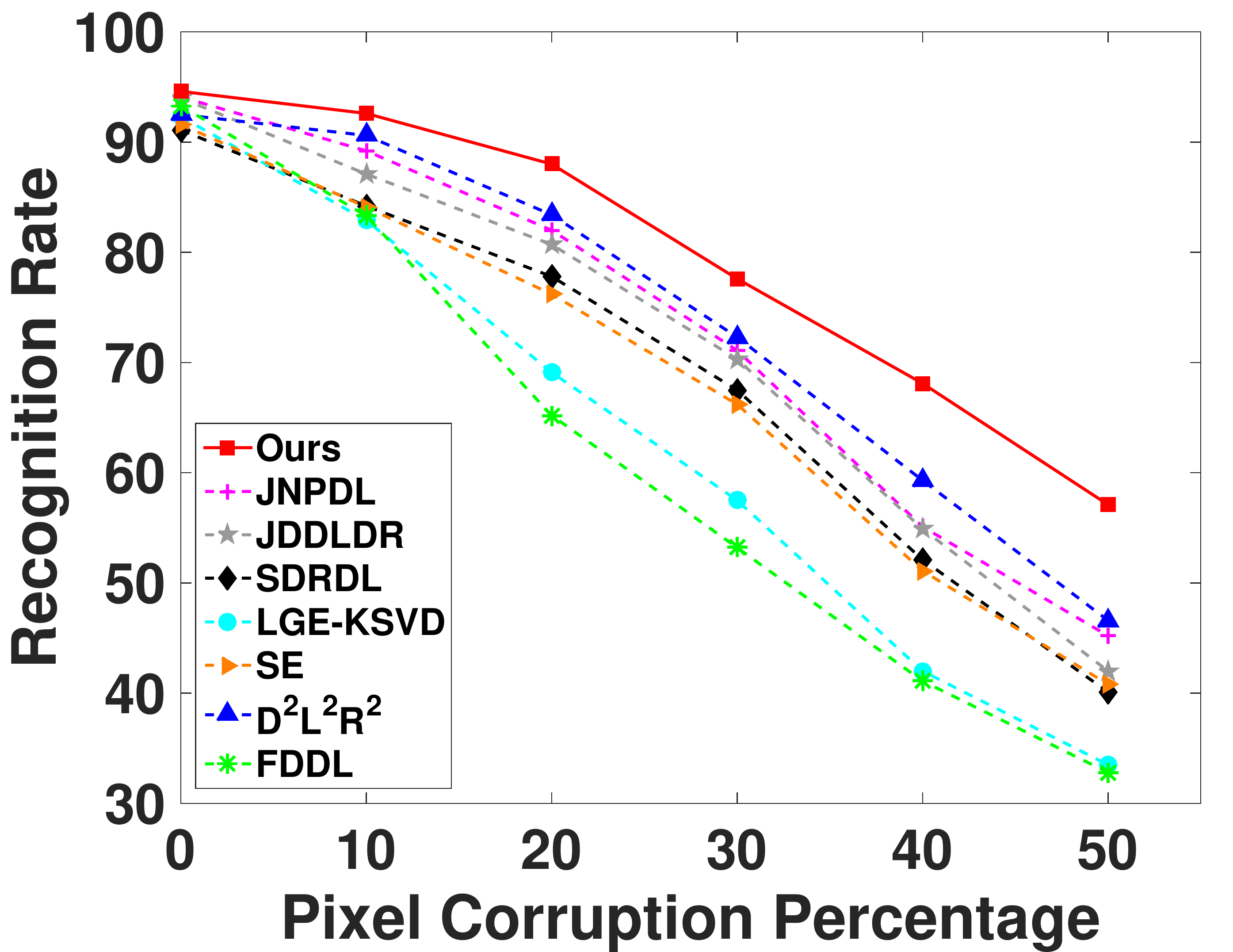} \label{fig:Yale-Pixel}}  
\hspace{2pt}
\subfloat[]{\includegraphics[width=4.2cm,keepaspectratio]{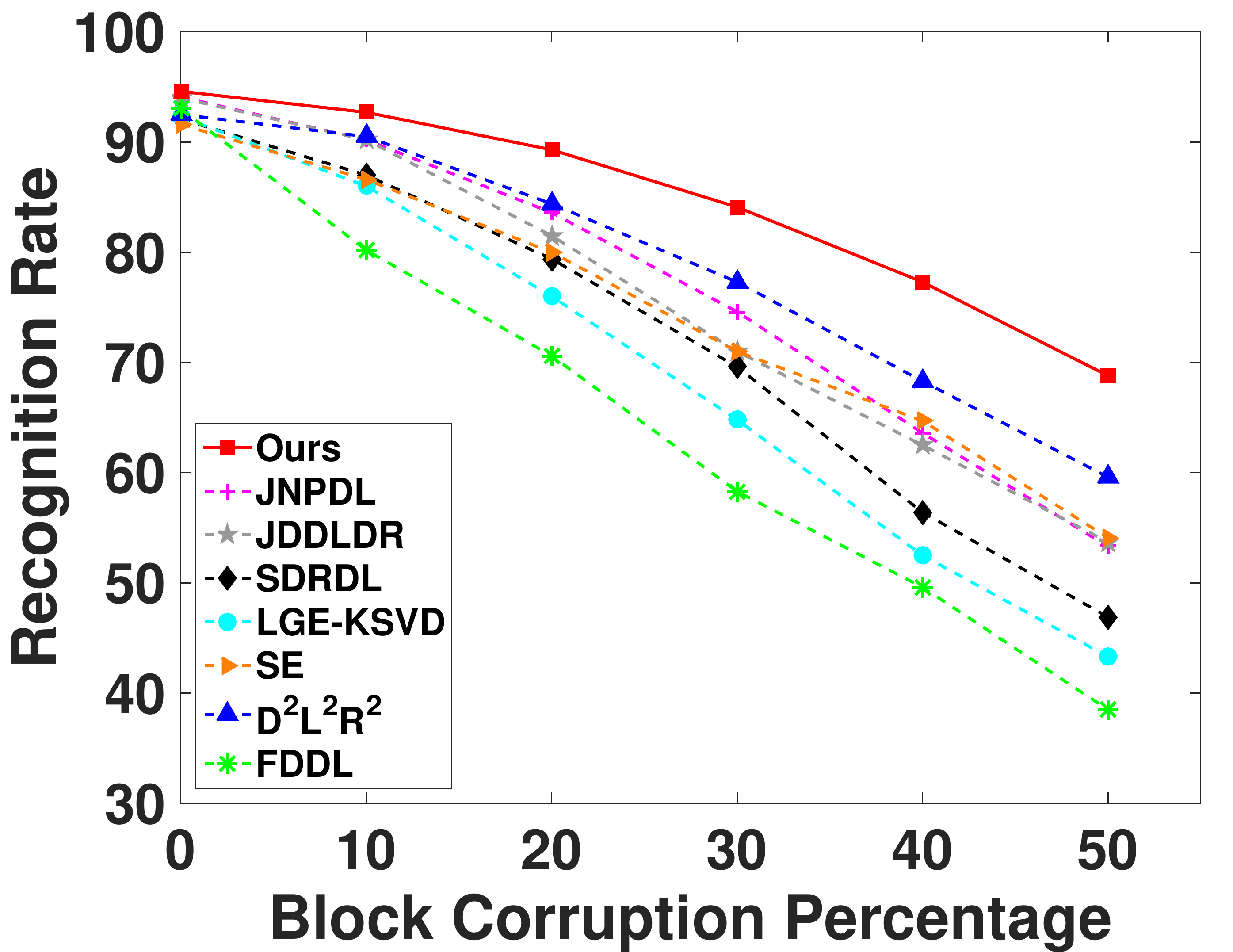} \label{fig:Yale-Block}}  
\vspace{0.2em}
\subfloat[]{\includegraphics[width=4.2cm,keepaspectratio]{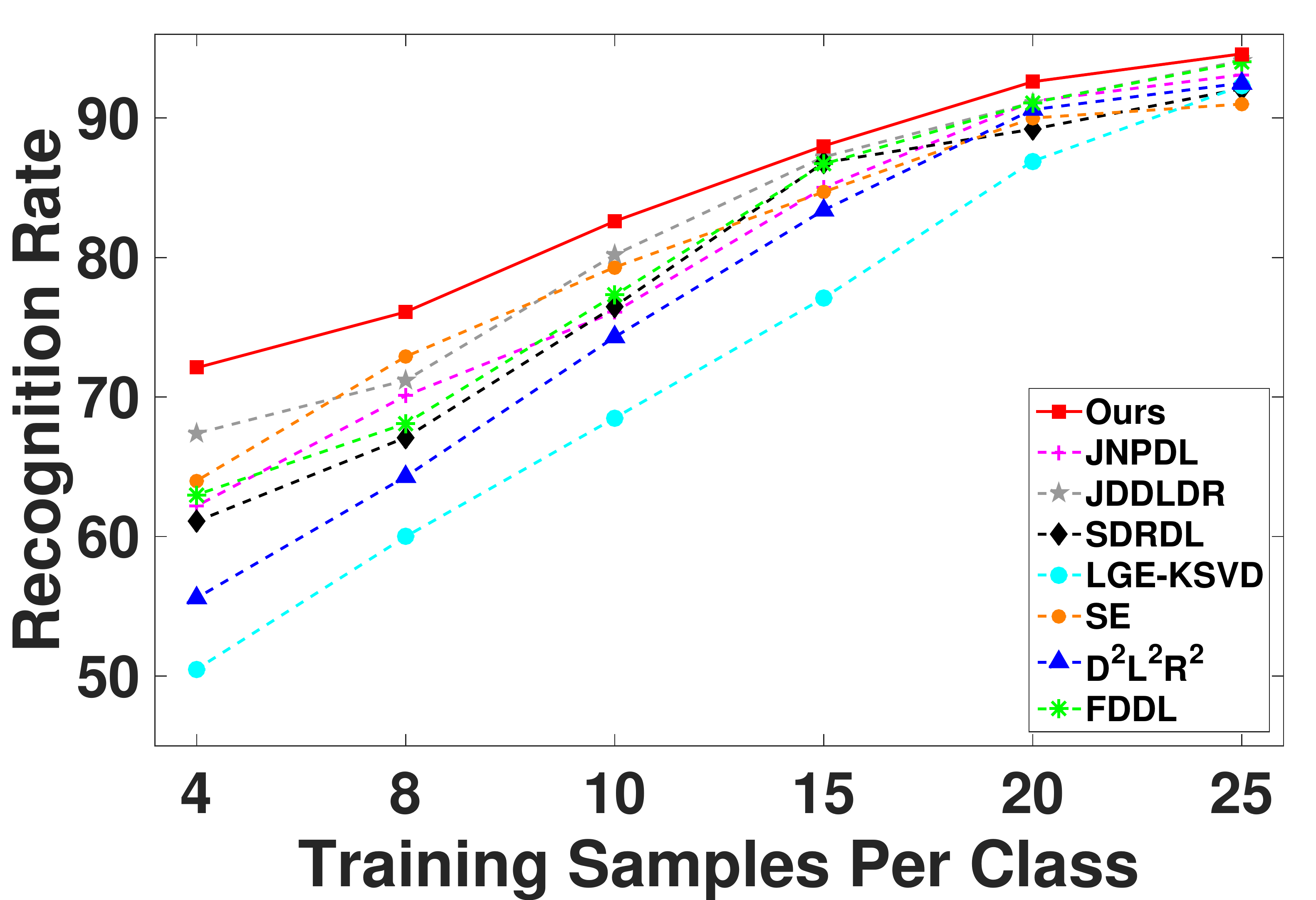}  \label{fig:Yale-TrSample}}  
\subfloat[]{\adjustbox{raise=1.5pc}{\includegraphics[width=4.1cm,keepaspectratio]{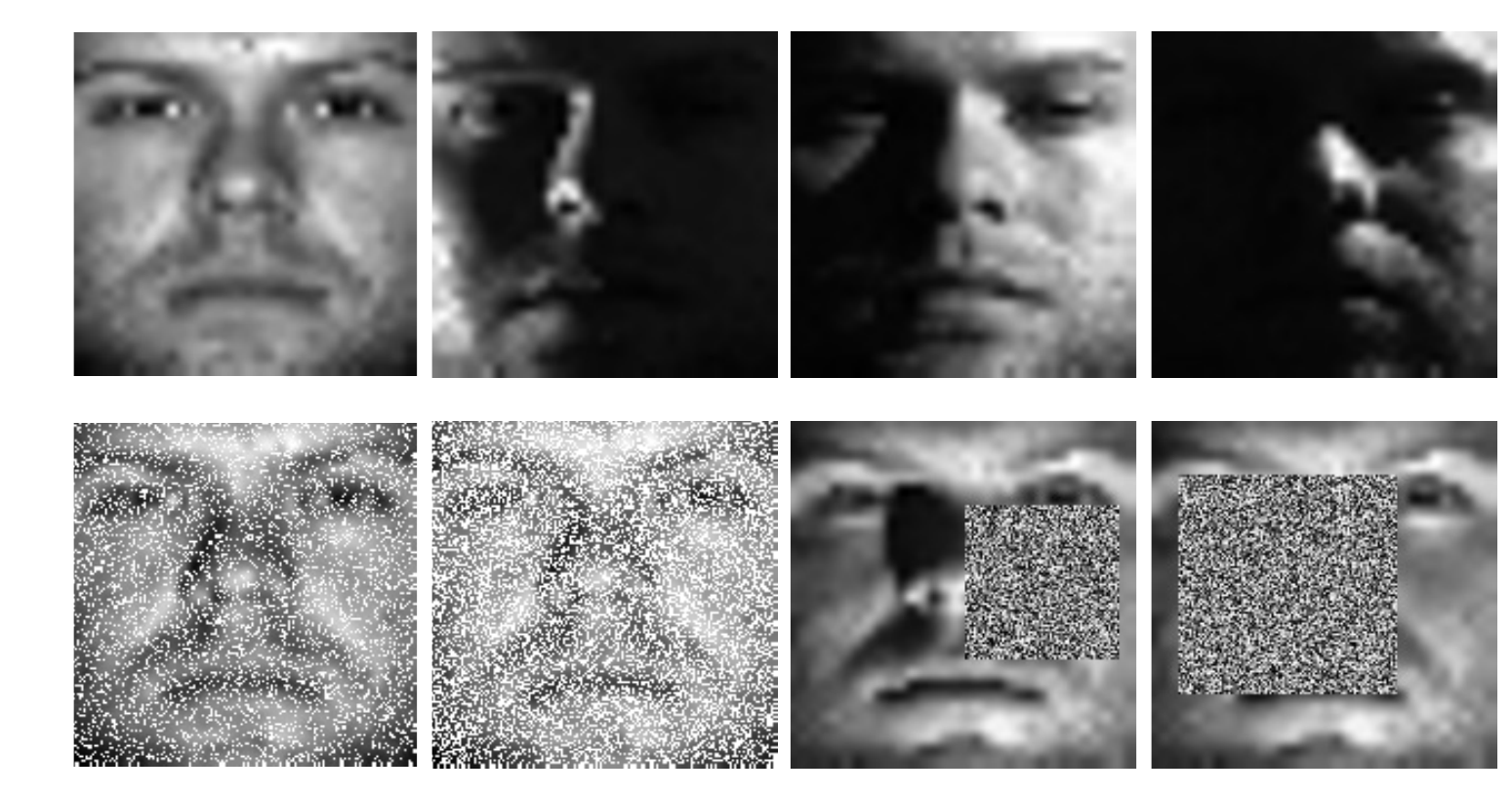} \label{fig:Faces-Yale}}} 
\caption{Recognition rates (\%) on the Extended YaleB dataset (a) with different levels of pixel corruption (b) with different levels of block corruption (c) with different number of training samples (d) samples of the Extended YaleB dataset; original, pixel corrupted (20\% and 40\%) and occluded (20\% and 40\%)}
\vspace{-1.5em}
\end{figure} 
%-------------------------------------------------------------------------------------------------------
%&&&&&&&&&&&&&&&&&&&&&&&&&&&&&&&&&&&&&&&&&&&&&&&&&&&&&&&&&&&&&&&&&&&&&&&&&&&&&&&&&&&&&&&&&&&&&&&&&&&&&&&&
\subsection{Face Recognition}
\textbf{Extended YaleB Dataset:} This dataset~\cite{Yale} contains $2,414$ frontal face images of $38$ human subjects captured under different illumination conditions. All the face images are cropped and resized to $55 \times 48$ and we randomly select $20$ images per class for training and the rest is used for test. To challenge ourselves, we also simulate various levels of corruption and occlusion. For pixel corruption, we replace a certain percentage (from $10\%$ to $50\%$) of randomly selected pixels of each image with pixel value $255$. For occlusion (block corruption), the images are manually corrupted by an unrelated block image at a random location and the percentage of corrupted area is increased from $10\%$ to $50\%$. Some of the original and corrupted/occluded images of this dataset can be seen in Figure~\ref{fig:Faces-Yale}. In the following experiments, all the methods utilize the raw images as the feature descriptor, except FDDL and D\textsuperscript{2}L\textsuperscript{2}R\textsuperscript{2} methods that initially use PCA to reduce the dimension of features, \ie the Eigenface is used as input.

We evaluate the robustness of our method to different levels of pixel and block corruption (from $10\%$ to $50\%$). For each level of corruption, the projected dimension varies between $5\%$ to $90\%$ of the original dimension ($2640$) and the best achieved result among all dimensions is reported. Figures~\ref{fig:Yale-Pixel} and~\ref{fig:Yale-Block} demonstrate that our method consistently obtains better performance than others in all levels of corruption. As the percentage of corruption/occlusion increases, the performance difference between JP-LRDL and other methods becomes more significant and this reflects the robustness of our method toward large noise. These figure also reflect that none of the existing joint DR-DL methods can achieve good performance for corrupted observations. Equally important, the best performance of JP-LRDL is achieved at $25\%$ of the original dimension, while that of existing joint DR-DL methods and DL methods occurs at $50\%$ and whole dimension, respectively. JP-LRDL is superior to other methods, even with fewer number of features.

Then, we randomly choose $4 \sim 25$ training samples per subject and evaluate the recognition rate on this dataset. Figure~\ref{fig:Yale-TrSample} shows that our results consistently outperform other counterparts under the same configurations. Moreover, significant improvements in the results are observed when there are few samples per subject. The proposed JP-LRDL is particularity less sensitive to small-sized dataset and maintains a relatively stable performance even in lower numbers. We also evaluate the running time of JP-LRDL and other competing methods on the Extended YaleB dataset in Figure~\ref{fig:Train-Time}. The training time is computed as the average over the entire training set, while fixing the projected dimension as $30\%$ of the original dimension. We used a machine with 12GB RAM and Intel Core i7-3770 CPU. Our method has a reasonable train time compared to other competing methods.
%&&&&&&&&&&&&&&&&&&&&&&&&&&&&&&&&&&&&&&&&&&&&&&&&&&&&&&&&&&&&&&&&&&&&&&&&&&&&&&&&&&&&&&&&&&&&&&&&&&&&&&&&
% --------------------------- AR recognition results -------------------------------------------------
\begin{figure}[b]
\centering
\subfloat[]{\includegraphics[width=4.2cm,keepaspectratio]{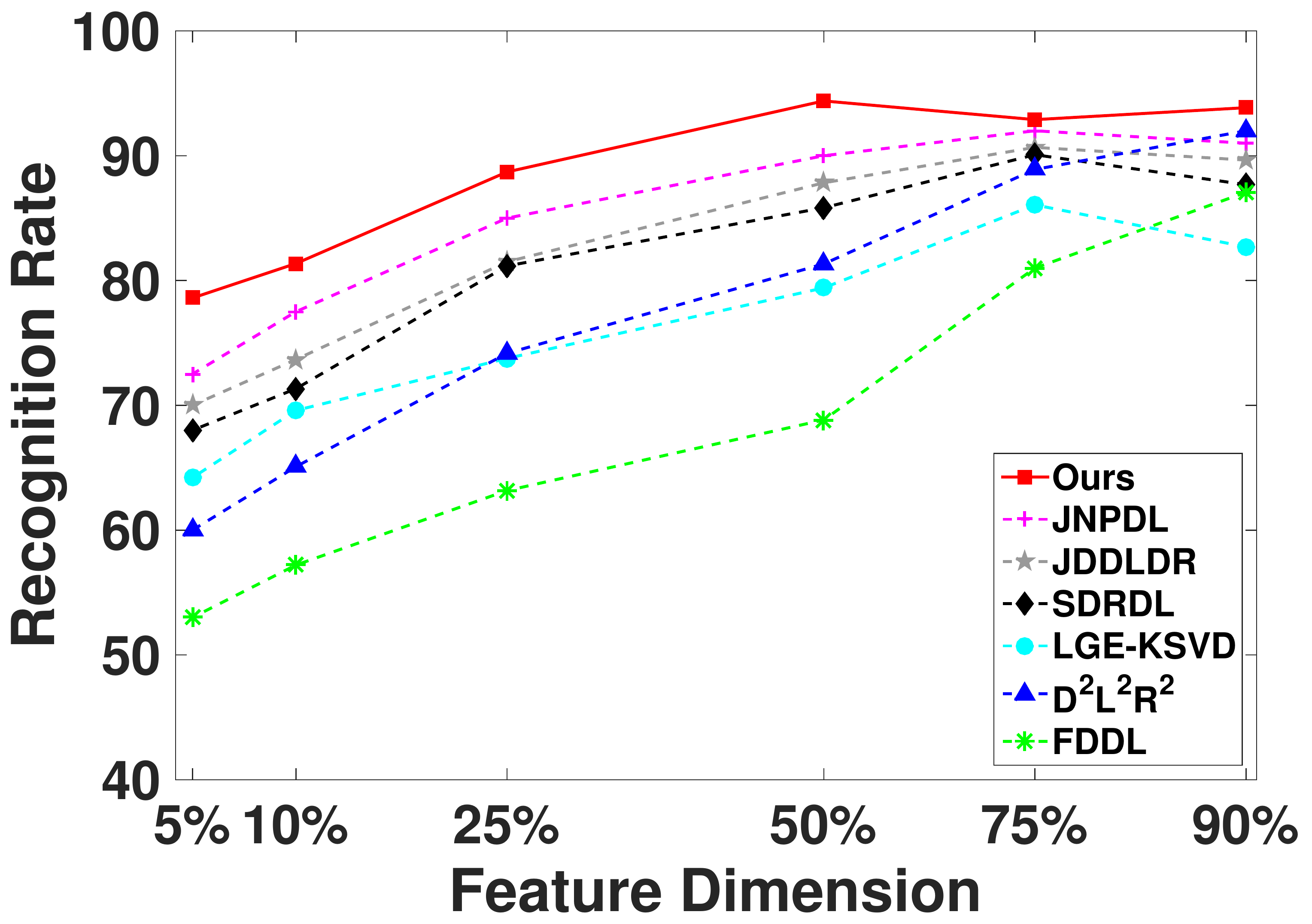} \label{fig:AR-Sunglasses}}  
\hspace{2pt}
\subfloat[]{\includegraphics[width=4.2cm,keepaspectratio]{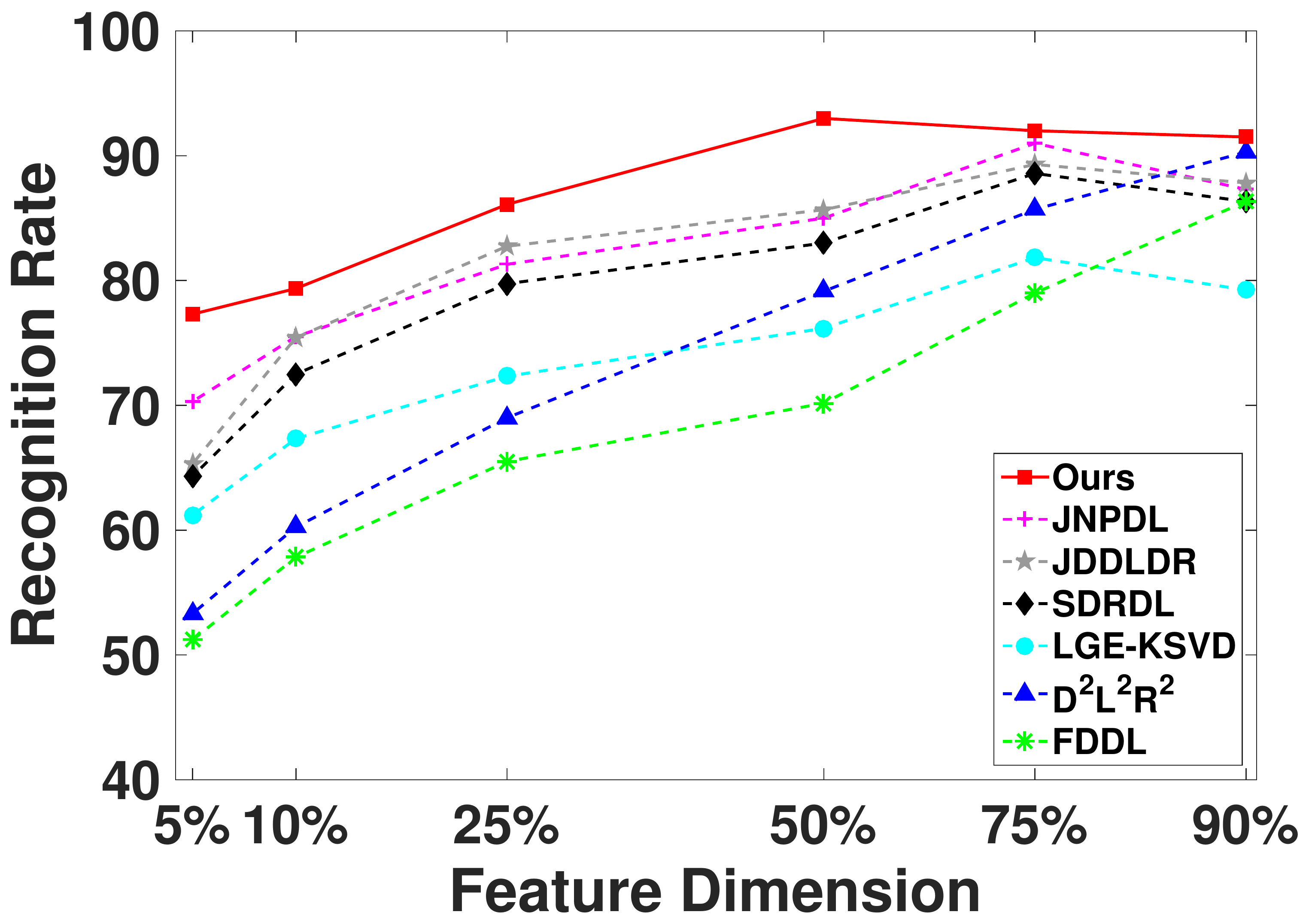} \label{fig:AR-Scarf}}  
\vspace{0.1em}
\subfloat[]{\includegraphics[width=4.2cm,keepaspectratio]{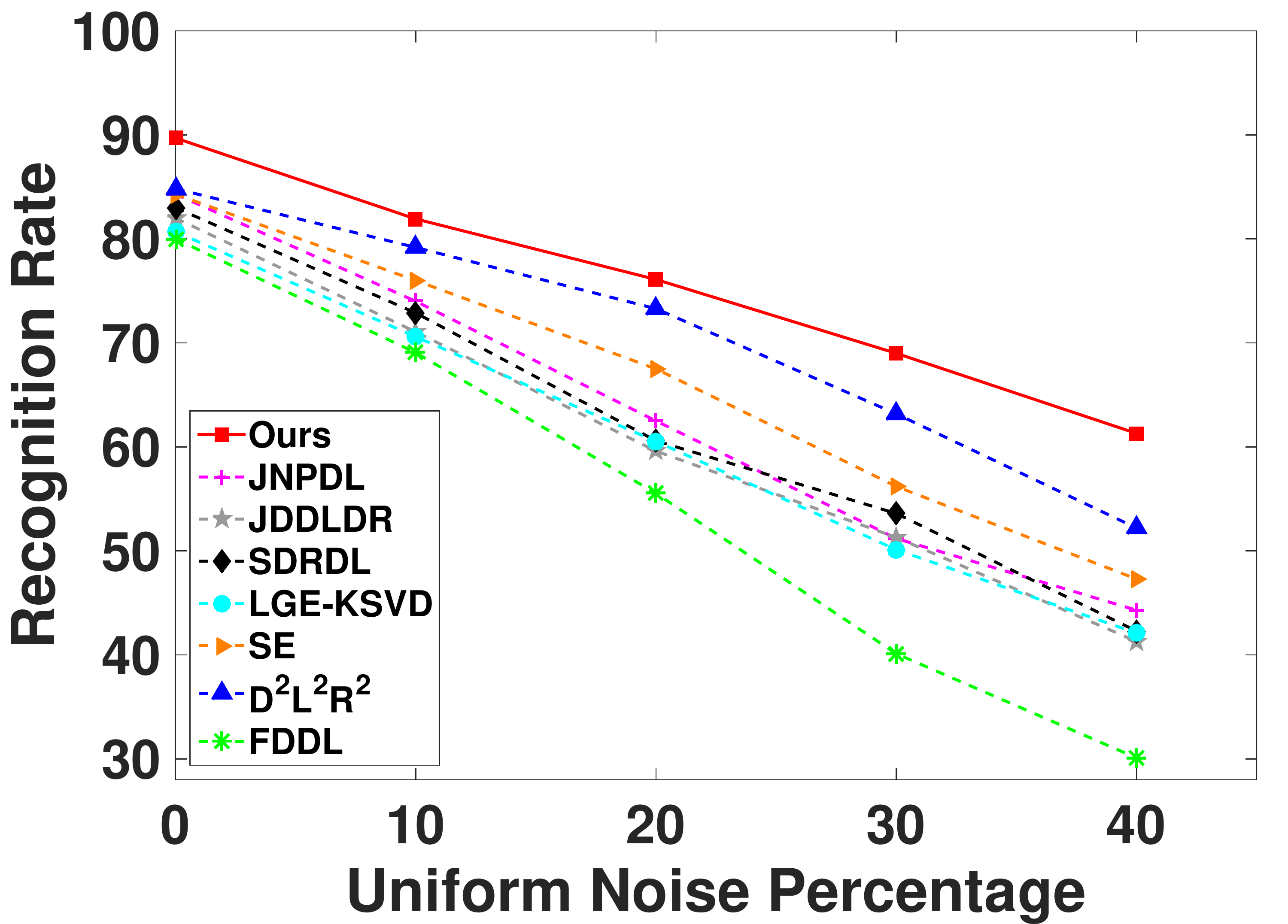}  \label{fig:AR-Mixed-Uniform}}  
\hspace{6pt}
\subfloat[]{\adjustbox{raise=1.2pc}{\includegraphics[width=3.8cm,keepaspectratio]{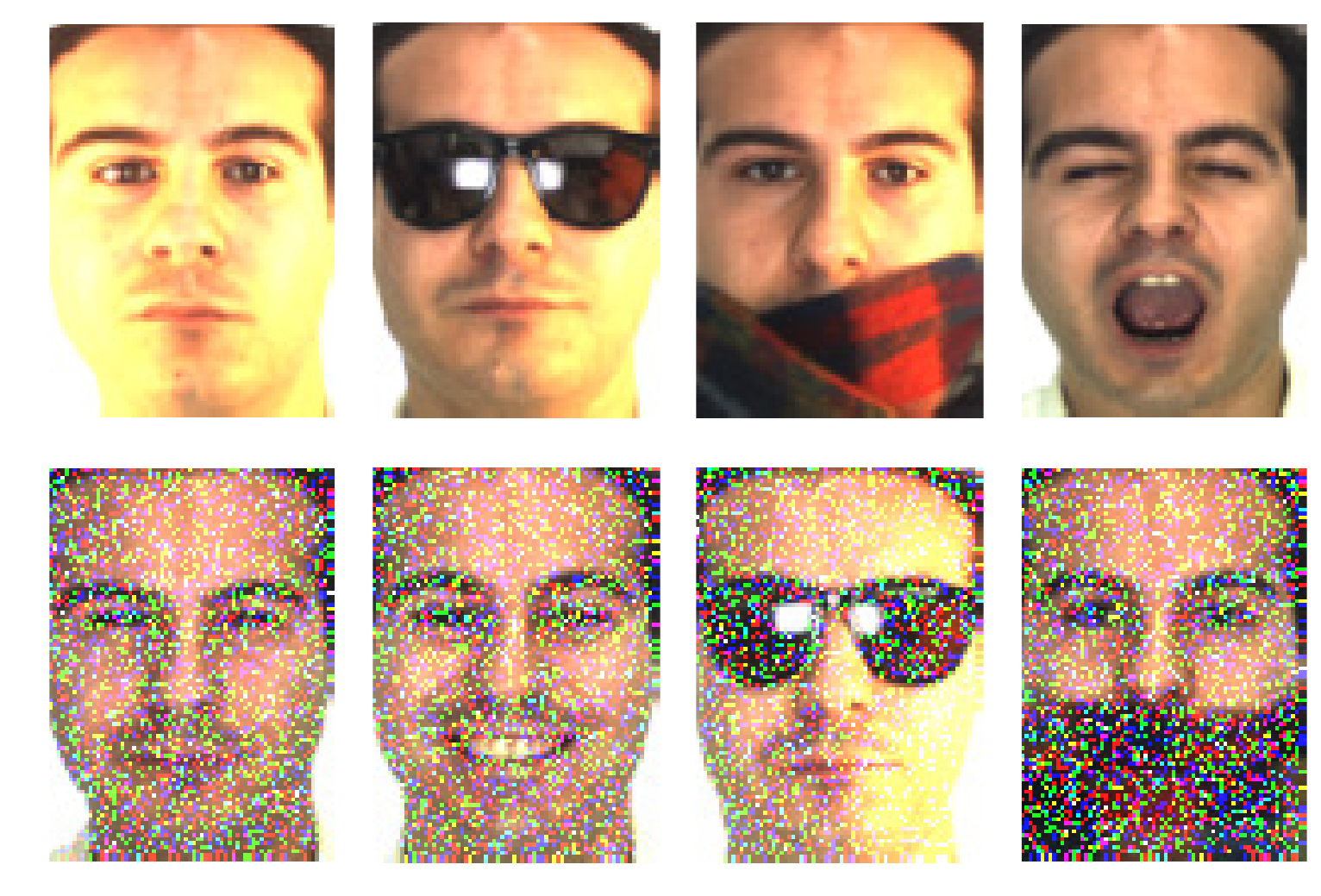} \label{fig:Faces-AR}}} 
\caption{Recognition rates (\%) on the AR dataset (a) versus varying feature dimension on Sunglasses scenario (b) versus varying feature dimension on Scarf scenario (c) with different levels of uniform noise on Mixed scenario (d) samples of the AR dataset; original and 20\% pixel corrupted}
\vspace{-1.5em}
\end{figure} 
%-------------------------------------------------------------------------------------------------------
%&&&&&&&&&&&&&&&&&&&&&&&&&&&&&&&&&&&&&&&&&&&&&&&&&&&&&&&&&&&&&&&&&&&&&&&&&&&&&&&&&&&&&&&&&&&&&&&&&&&&&&&&

\vspace{1em}
\textbf{AR Dataset:} The AR face dataset~\cite{AR} includes over $4,000$ frontal face images from $126$ individuals. We select a subset of $2,600$ images from $50$ male and $50$ female subjects in the experiments. These images include different facial expressions, illumination conditions and disguises. In each session, each person has $13$ images, of which $3$ are obscured by scarves, $3$ by sunglasses and the remaining ones are of different facial expressions or illumination variations which we refer to as unobscured images. Each face image is resized to $55 \times 40$ and following the protocol in~\cite{Structured-LR-DL}, experiments are conducted under three different scenarios:

$-$\textbf{Sunglasses:} We select $7$ unobscured images and $1$ image with sunglasses from the first session as training samples for each person. The rest of unobscured images from the second session and the rest of images with sunglasses are used for testing. Sunglasses occlude about $20\%$ of images.

$-$\textbf{Scarf:} We choose $8$ training images ($7$ unobscured and $1$ with scarf) from the first session for training, and $12$ test images including $7$ unobscured images from the second session and the remaining $5$ images with scarf from two sessions for testing. The scarf covers around $40\%$ images.

$-$\textbf{Mixed:} We consider the case in which both training and test images are occluded by sunglasses and scarf. We select $7$ unobscured, plus $2$ occluded images ($1$ with sunglasses, $1$ by scarf) from the first session for training and the remaining $17$ images in two sessions for testing per class.

In the following experiments, we use the raw images as the feature descriptor for all the methods, except FDDL and D\textsuperscript{2}L\textsuperscript{2}R\textsuperscript{2}, which use Randomface~\cite{LC-KSVD} that is generated by projecting a face image onto a random vector. 
First, we evaluate the robustness of our method in small-sized, large intra-class variability datasets. We consider Sunglasses and Scarf scenarios and to have more challenge, all the training images are manually corrupted by $20\%$ pixel corruption. Then we vary the feature dimension from $5\%$ to $90\%$ of the original dimension ($2200$) and report the recognition rate. Figure~\ref{fig:Faces-AR} shows some of these original and pixel corrupted images. Figures~\ref{fig:AR-Sunglasses},\ref{fig:AR-Scarf} show the recognition rates of JP-LRDL and competing methods over these two scenarios. Our approach achieves the best results compared to the competing methods, across all dimensions and maintains a relatively stable performance in lower dimensions. JP-LRDL is able to achieve the best recognition rate while using $50\%$ of all features. We also not that existing joint DR-DL methods perform better than DL methods in lower dimensions due to the learned projection matrix, which is reasonably more powerful than random projection. 

Then, we evaluate our algorithm on the Mixed scenario and to challenge ourselves, we also simulate uniform noise, such that a percentage of randomly chosen pixels of each image, are replaced with samples from a uniform distribution over $[0; V_{max}]$, where $V_{max}$ is the largest possible pixel value in the image. In this experiment, the projected dimension is fixed as $30\%$ of the original dimension and the recognition accuracy under different levels of corruption is reported in Figure~\ref{fig:AR-Mixed-Uniform}. One may infer JP-LRDL shows robustness to occlusions, severe corruption, illumination and expression changes; however, the existing methods fail to handle these variations. Furthermore, JP-LRDL is able preserve the discriminative information, even in relatively low dimensions.
%&&&&&&&&&&&&&&&&&&&&&&&&&&&&&&&&&&&&&&&&&&&&&&&&&&&&&&&&&&&&&&&&&&&&&&&&&&&&&&&&&&&&&&&&&&&&&&&&&&&&&&&&
% --------------------------- Images of LFW ------------------------------------------------------------
\begin{figure}[t]
\centering
{\includegraphics[width=7.5cm,keepaspectratio]{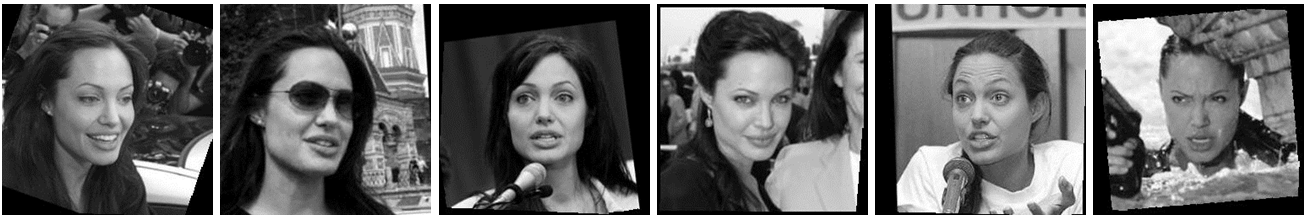} \label{fig:Faces-LFW}}  
\caption{Sample faces images of the LFWa dataset}
\vspace{-1.5em}
\end{figure} 
%-------------------------------------------------------------------------------------------------------
%&&&&&&&&&&&&&&&&&&&&&&&&&&&&&&&&&&&&&&&&&&&&&&&&&&&&&&&&&&&&&&&&&&&&&&&&&&&&&&&&&&&&&&&&&&&&&&&&&&&&&&&&
%&&&&&&&&&&&&&&&&&&&&&&&&&&&&&&&&&&&&&&&&&&&&&&&&&&&&&&&&&&&&&&&&&&&&&&&&&&&&&&&&&&&&&&&&&&&&&&&&&&&&&&&&
% --------------------------- LFW recognition results ---------------------------------------------------
\begin{table}[b]
\caption{Recognition rates (\%) on the LFWa dataset}
\label{table:LFW}
\centering
\begin{tabular}{|l|l|l|l|}
\hline
Method & Rec. Rate (\%) & Method & Rec. Rate (\%) \Tstrut\Bstrut\\
\hline \hline 
JNPDL~\cite{JNPDL}     & 78.10 &   JDDLDR~\cite{JDDRDL}      & 72.40 \Tstrut\Bstrut\\
SDRDL~\cite{Simul-DL}  & 71.25 &   LGE-KSVD~\cite{LGE-KSVD}  & 70.42 \Tstrut\Bstrut\\
D\textsuperscript{2}L\textsuperscript{2}R\textsuperscript{2}~\cite{D2L2R2} & 75.20 & FDDL~\cite{FDDL} & 74.81 \Tstrut\Bstrut\\
SE~\cite{SE}           & 76.34 &   Our method & 79.87 \Tstrut\Bstrut\\
\hline
\end{tabular}
\end{table}
%-------------------------------------------------------------------------------------------------------
%&&&&&&&&&&&&&&&&&&&&&&&&&&&&&&&&&&&&&&&&&&&&&&&&&&&&&&&&&&&&&&&&&&&&&&&&&&&&&&&&&&&&&&&&&&&&&&&&&&&&&&&&

\vspace{1em}
\textbf{LFW Dataset:} Besides tests with laboratory face datasets, we also evaluate the JP-LRDL on the LFW dataset~\cite{LFW} for unconstrained face verification. LFW contains $13,233$ face images of $5,749$ different individuals, collected from the web with large variations in pose, expression, illumination, clothing, hairstyles, occlusion, etc. Here, we use LWFa dataset~\cite{LFW-a}, which is an aligned version of LFW. We use $143$ subject with no less than $11$ samples per subject in LFWa dataset ($4174$ images in total) to perform the experiment. The first $10$ samples are selected as the training samples and the rest is for testing. Face images are cropped and normalized to the size of $60 \times 54$ and the projected dimension is set as $1000$. Also, PCA is used for DR of FDDL and D\textsuperscript{2}L\textsuperscript{2}R\textsuperscript{2} methods. Table~\ref{table:LFW} lists the recognition rates of all the methods, and similar to previous results, JP-LRDL achieves the best performance. These results confirm that the proposed method not only effectively learn robust feature representations in controlled scenarios, but also have excellent discrimination ability for face images that collected in uncontrolled conditions and have high variation.
%&&&&&&&&&&&&&&&&&&&&&&&&&&&&&&&&&&&&&&&&&&&&&&&&&&&&&&&&&&&&&&&&&&&&&&&&&&&&&&&&&&&&&&&&&&&&&&&&&&&&&&&&
% --------------------------- Images of COIL and Caltech --------------------------------------------------
\begin{figure}[t]
\centering
\subfloat[]{\includegraphics[width=4.2cm,keepaspectratio]{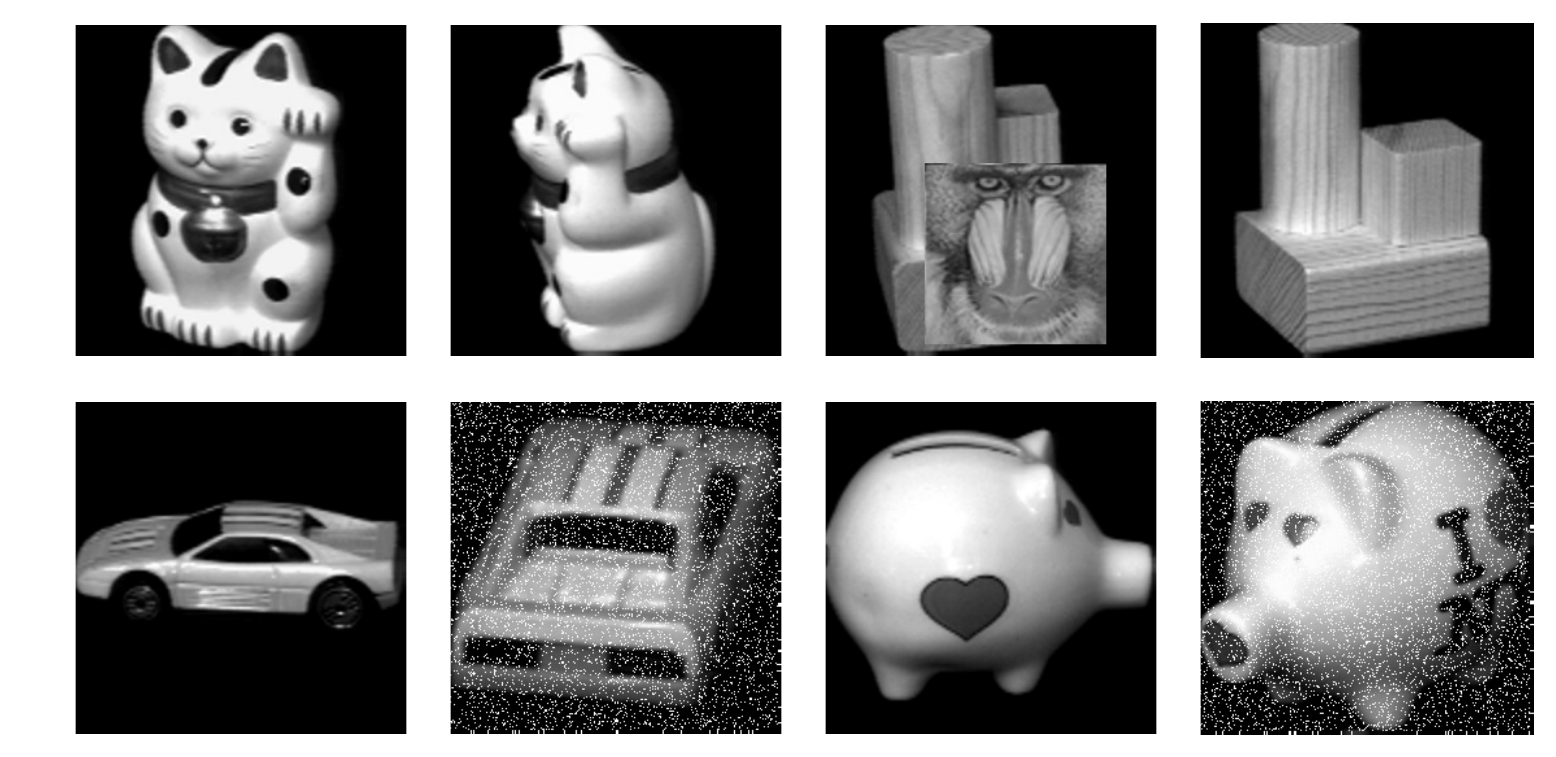} \label{fig:Objects-COIL}}  
\hspace{2pt}
\subfloat[]{\includegraphics[width=4.2cm,keepaspectratio]{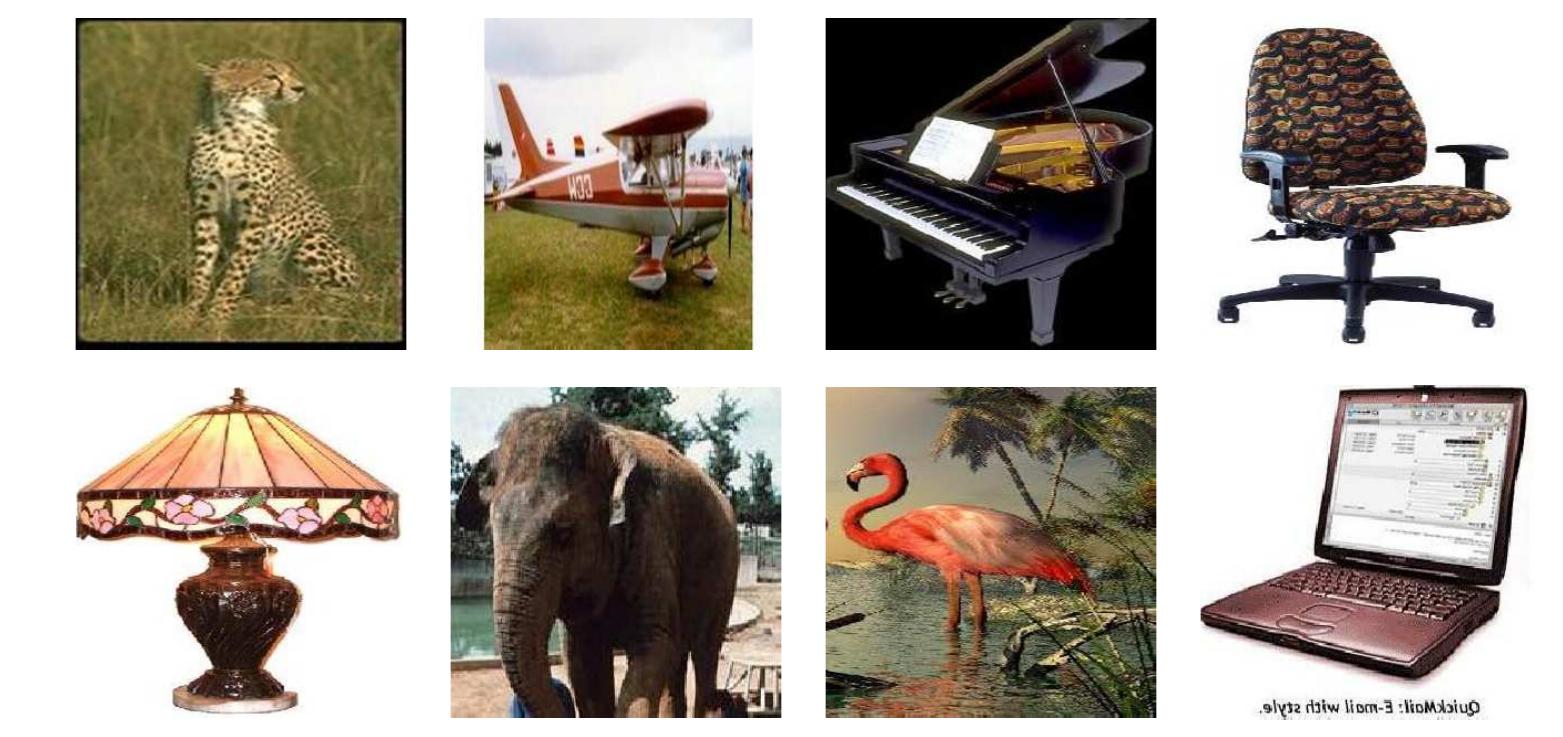} \label{fig:Objects-Caltech}}
\caption{Sample images of (a) the COIL dataset; original, $10\%$ pixel corrupted and 30\% occluded images with unrelated block (b) the Caltech-101 dataset}
\vspace{-1.5em}
\end{figure} 
%-------------------------------------------------------------------------------------------------------
%&&&&&&&&&&&&&&&&&&&&&&&&&&&&&&&&&&&&&&&&&&&&&&&&&&&&&&&&&&&&&&&&&&&&&&&&&&&&&&&&&&&&&&&&&&&&&&&&&&&&&&&&
\subsection{Object Recognition}
\textbf{COIL Dataset:} The COIL dataset~\cite{COIL} contains various views of $100$ objects with different lighting conditions and scales. In our experiments, the images are resized to $32 \times 32$ and the training set is constructed by randomly selecting $10$ images per object from available $72$ images. Some of the original and corrupted images can be found in Figure~\ref{fig:Objects-COIL}.

\begin{figure}[t]
\centering
\subfloat[]{\includegraphics[width=4.2cm,keepaspectratio]{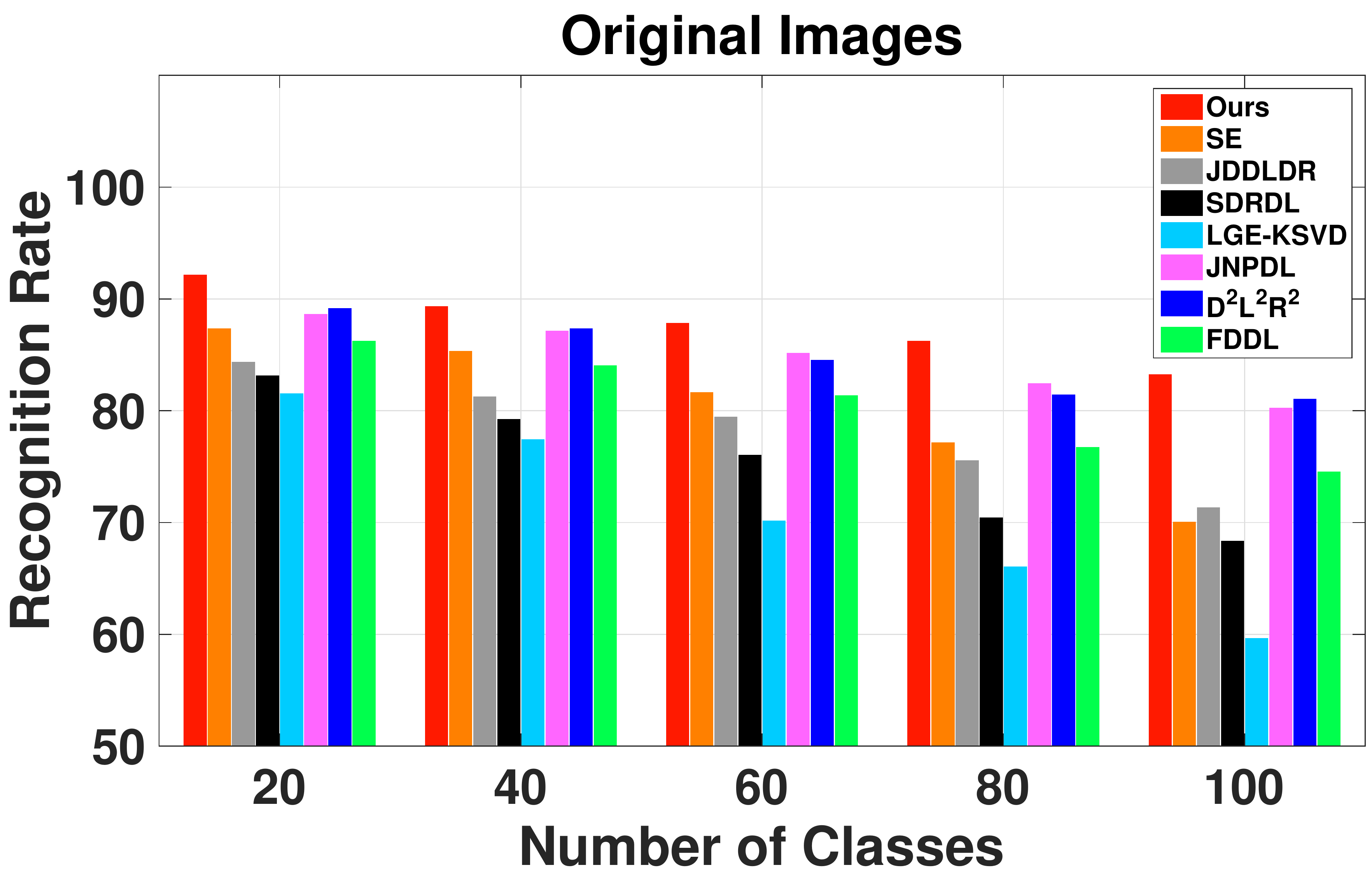} \label{fig:Coil-Org}}  
\hspace{2pt}
\subfloat[]{\includegraphics[width=4.2cm,keepaspectratio]{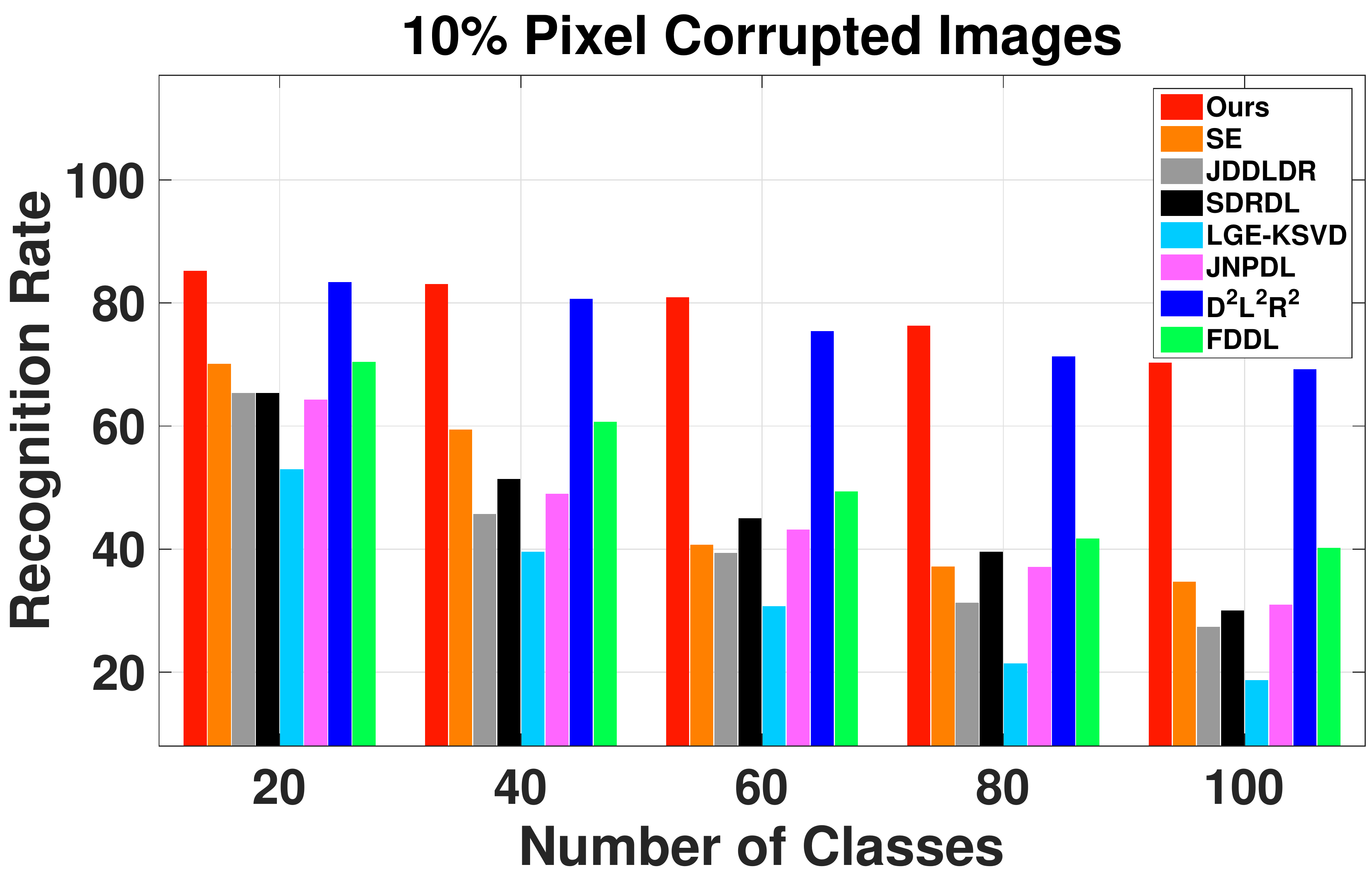} \label{fig:Coil-Noise}}  
\vspace{0.2em}
\subfloat[]{\includegraphics[width=4.2cm,keepaspectratio]{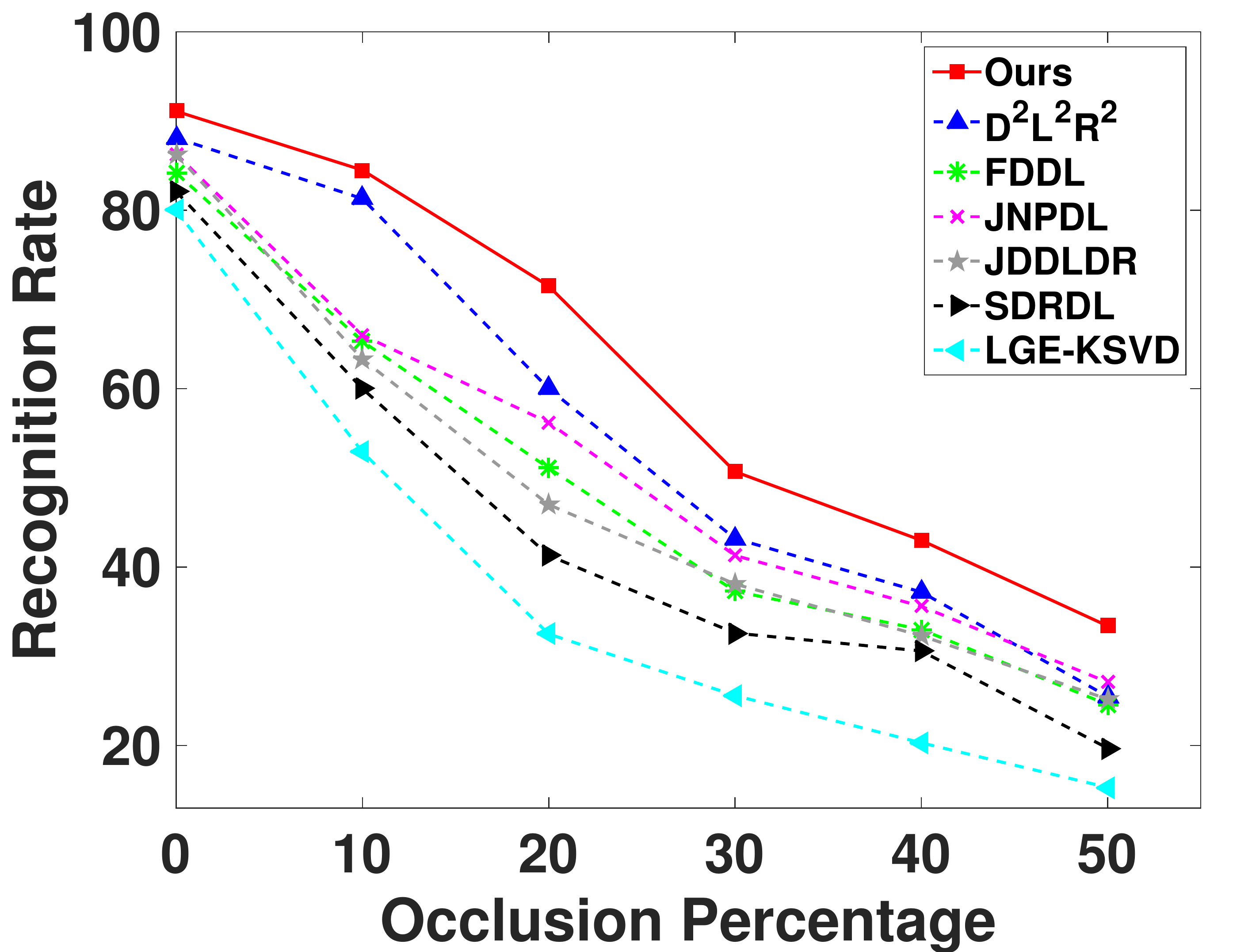} \label{fig:Coil20-Noise-Dim}}  
\hspace{2pt}
\subfloat[]{\adjustbox{raise=0.5pc}{\includegraphics[width=4.2cm,keepaspectratio]{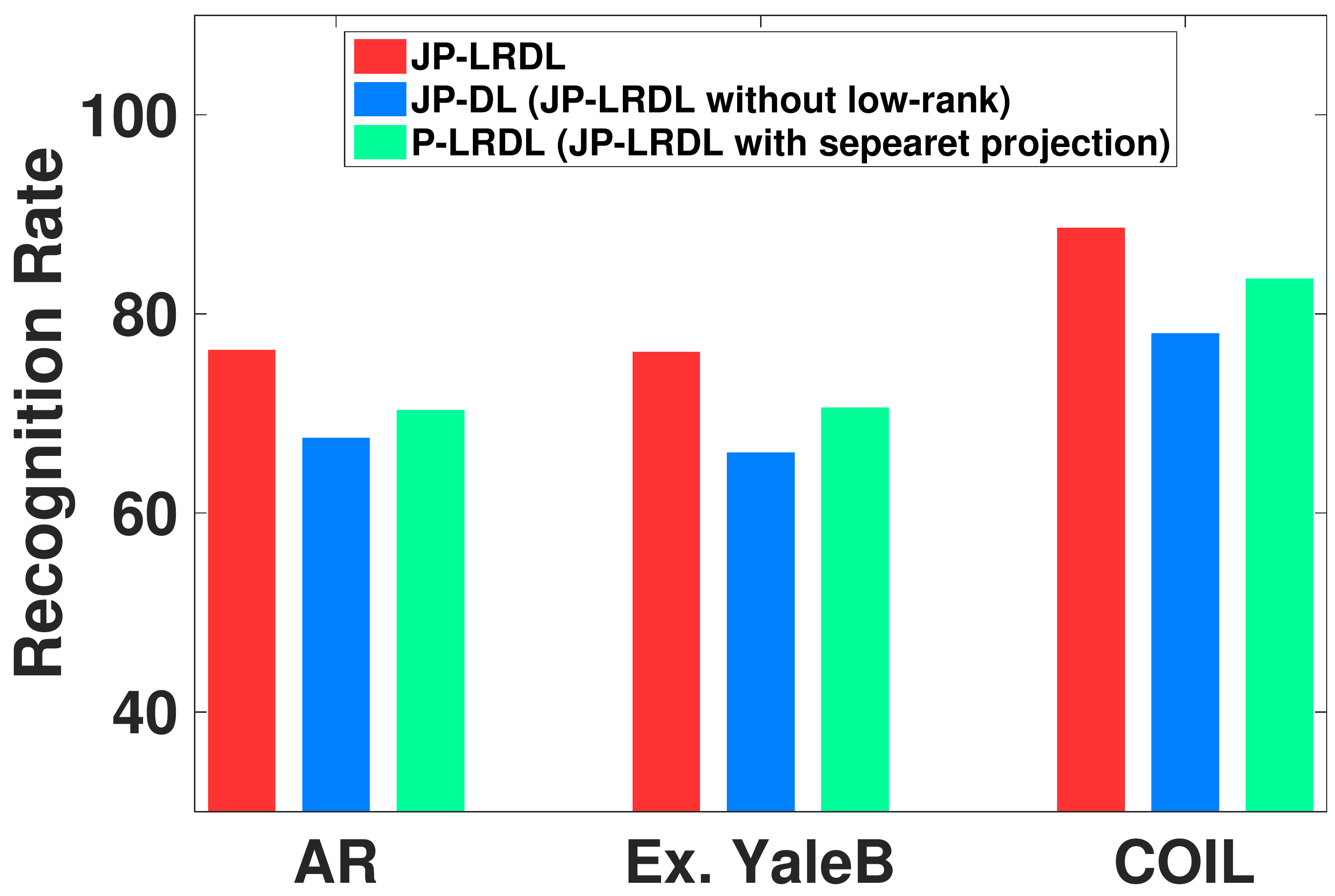} \label{fig:Components}}}  
\caption{Recognition rates (\%) on the COIL dataset with (a) original images, (b) 10\% pixel corrupted images, and (c) versus different levels of occlusion on COIL-20 dataset (d) The role of different components of JP-LRDL on several datasets}
\vspace{-1.5em}
\end{figure}

We evaluate the scalability of our method and the competing methods by increasing the number of objects (\ie classes) from $10$ to $100$. In addition to alternative viewpoints, we also test the robustness of different methods to simulated noise by adding $10\%$ pixel corruption to the original images.
Figure~\ref{fig:Coil-Org},~\ref{fig:Coil-Noise} show the average recognition rates for all compared methods over original images and $10\%$ pixel corrupted images for different class numbers, respectively. Like before, for all the methods, the projected dimension is varied from $5\%$ to $90\%$ of the original dimension ($1024$) and the best achieved performance is reported. 
It can be observed that the proposed JP-LRDL outperforms the competing methods and the difference becomes more meaningful, when data are contaminated with simulated noise. All the methods, except D\textsuperscript{2}L\textsuperscript{2}R\textsuperscript{2} and our approach, which utilize LR constraint, have difficulty obtaining reasonable results for corrupted data. In particular, our method achieves remarkable performance and demonstrates good scalability in both scenarios.

Moreover, we simulate various levels of contiguous occlusion (from 10\% to 50\%), by replacing a randomly located square block of each test image of COIL-20 dataset with an unrelated image. We also set the feature dimension as $30\%$ of the original dimension and the average recognition rates are illustrated in Figure~\ref{fig:Coil20-Noise-Dim}. JP-LRDL achieves the highest recognition rate under different levels of occlusion. 
%&&&&&&&&&&&&&&&&&&&&&&&&&&&&&&&&&&&&&&&&&&&&&&&&&&&&&&&&&&&&&&&&&&&&&&&&&&&&&&&&&&&&&&&&&&&&&&&&&&&&&&&&
% --------------------------- COIL recognition results ------------------------------------------------
%-------------------------------------------------------------------------------------------------------
%&&&&&&&&&&&&&&&&&&&&&&&&&&&&&&&&&&&&&&&&&&&&&&&&&&&&&&&&&&&&&&&&&&&&&&&&&&&&&&&&&&&&&&&&&&&&&&&&&&&&&&&&

Finally, we design an experiment to show the efficiency of different components of the proposed JP-LRDL framework. To verify the efficacy of LR constraint in the framework, we remove $\lambda_3 \sum_{\substack{i=1}}^K \norm[]{D_i}_{*}$ from Equation~\eqref{eq1}. In similar fashion, to evaluate the importance of joint DR and DL process, we remove the projection learning part from JP-LRDL, which means that the projection matrix and structured dictionary are learned from training samples separately. We call these two strategies JP-DL and P-LRDL respectively and compare them with the proposed JP-LRDL on three datasets in Figure~\ref{fig:Components}. In these experiments, the projected dimension methods is set to $10\%$ of the original dimension and the images are corrupted by $20\%$ block occlusion. For the AR and Extended YaleB datasets, we follow the Mixed scenario and regular experiment protocols, respectively. For the COIL dataset, we utilize first $20$ classes. According to the results, once the LR regularization is removed, the recognition rate drops significantly in all datasets. Also, we note that JP-LRDL outperforms P-LRDL (with separate projection) and this is mainly due to the fact that some useful information for DL maybe lost in the separate projection learning phase. The joint learning framework enhances the classification performance, especially when data are highly contaminated and dimension is relatively low.
%&&&&&&&&&&&&&&&&&&&&&&&&&&&&&&&&&&&&&&&&&&&&&&&&&&&&&&&&&&&&&&&&&&&&&&&&&&&&&&&&&&&&&&&&&&&&&&&&&&&&&&&&
% --------------------------- Results on Caltech tr samples ---------------------------------------------
\begin{figure}[t]
\centering
\subfloat[]{\includegraphics[width=4.2cm,keepaspectratio]{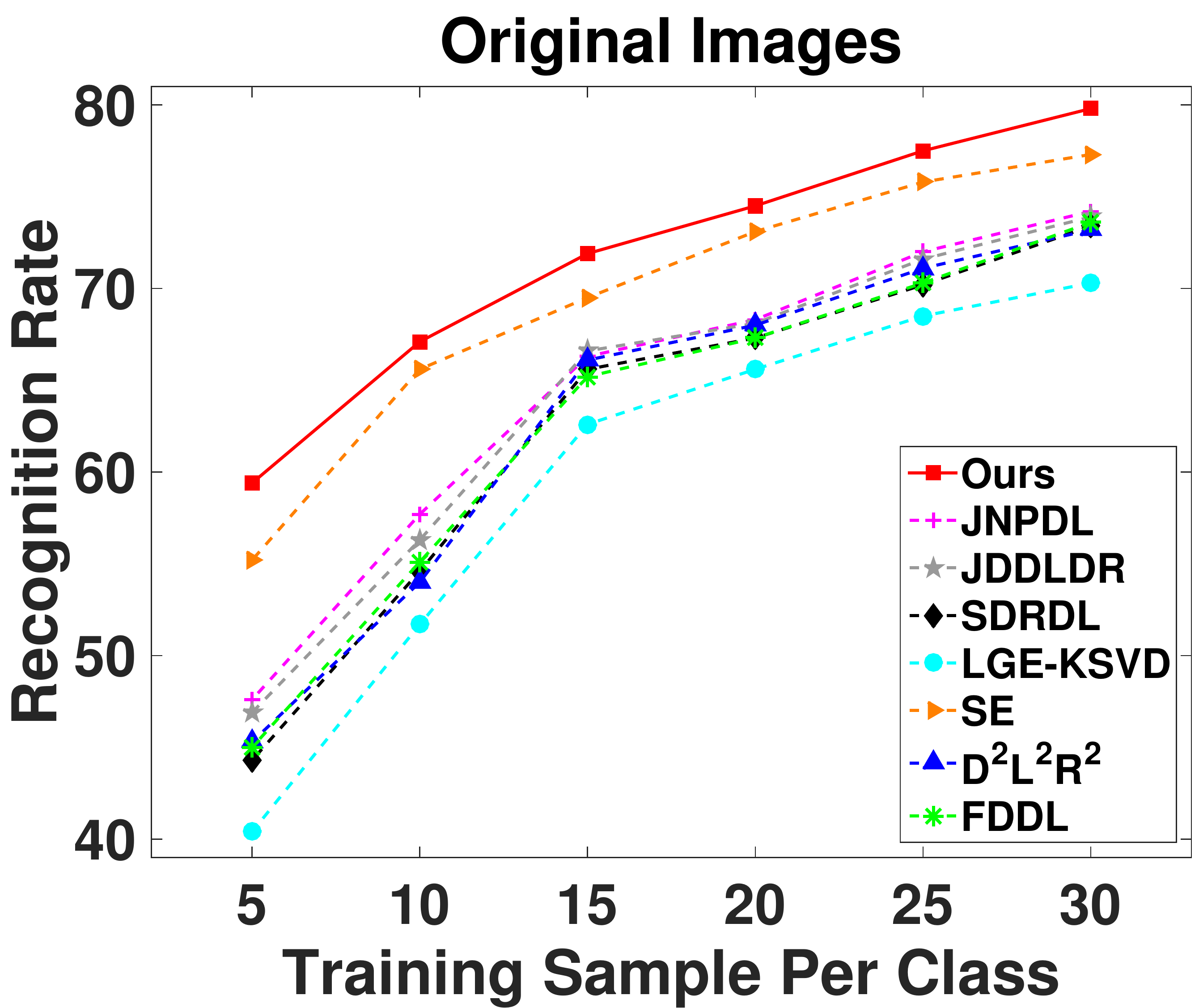} \label{fig:Caltech-Org-Train-Size}}  
\hspace{2pt}
\subfloat[]{\includegraphics[width=4.2cm,keepaspectratio]{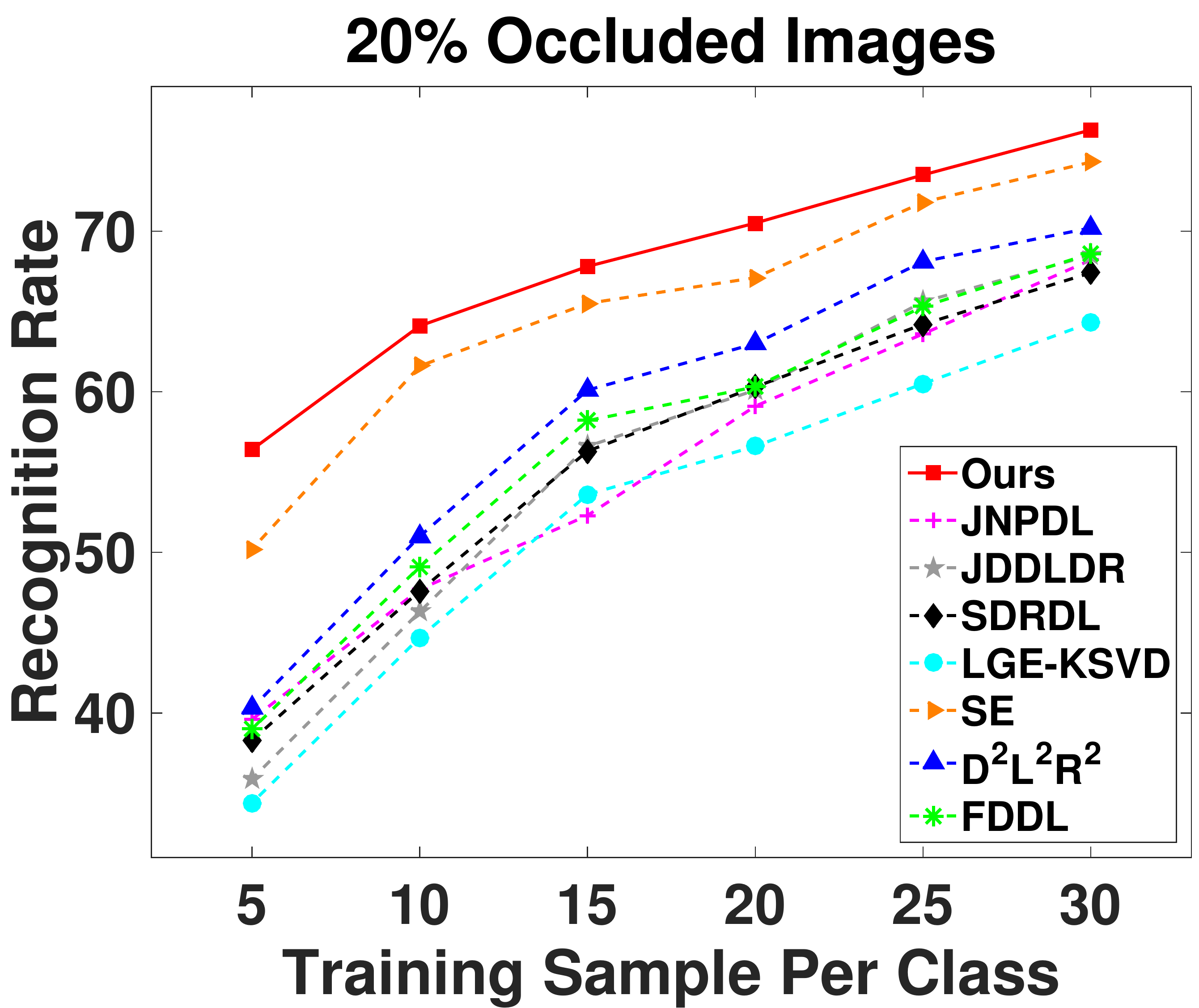} \label{fig:Caltech-Noise-Train-Size}}  
\caption{Recognition rates (\%) on the Caltech-101 dataset with different number of training samples on (a) original images (b) 20\% occluded images}
\vspace{-1.5em}
\end{figure} 
%-------------------------------------------------------------------------------------------------------
%&&&&&&&&&&&&&&&&&&&&&&&&&&&&&&&&&&&&&&&&&&&&&&&&&&&&&&&&&&&&&&&&&&&&&&&&&&&&&&&&&&&&&&&&&&&&&&&&&&&&&&&&

\vspace{1em}
\textbf{Caltech-101 Dataset:} The Caltech-101 database~\cite{Caltech} contains over $9000$ images from $101$ different object categories such as animals, flowers, trees, etc., and $1$ background class. The number of images in each class is greatly unbalanced, varying from $31$ to $800$. Figure~\ref{fig:Objects-Caltech} shows some sample images from this dataset. We evaluate our method using dense SIFT-based spatial pyramid features~\cite{LC-KSVD} and set the projected dimension as $3000$. We run the experiments with $15$ and $30$ randomly chosen training images per category and this process is repeated $10$ times with different random spits of the training and testing images to obtain reliable results. The final recognition rates are reported as the average of each run and summarized in Table~\ref{table:Caltech}.
%&&&&&&&&&&&&&&&&&&&&&&&&&&&&&&&&&&&&&&&&&&&&&&&&&&&&&&&&&&&&&&&&&&&&&&&&&&&&&&&&&&&&&&&&&&&&&&&&&&&&&&&&
% --------------------------- Caltech recognition results -----------------------------------------------
\begin{table}[b]
\caption{Recognition rates (\%) on the Caltech-101 dataset}
\label{table:Caltech}
\centering
\begin{tabular}{|l|c|c|}
\hline
Number of Training Samples & $15$ & $30$ \Tstrut\Bstrut\\
\hline \hline 
JNPDL~\cite{JNPDL}       & 66.83 & 74.61 \Tstrut\Bstrut\\
JDDLDR~\cite{JDDRDL}     & 67.70 & 73.90 \Tstrut\Bstrut\\
SDRDL~\cite{Simul-DL}    & 65.62 & 73.25 \Tstrut\Bstrut\\
LGE-KSVD~\cite{LGE-KSVD} & 62.23 & 70.42 \Tstrut\Bstrut\\
SE~\cite{SE}             & 69.50 & 77.34 \Tstrut\Bstrut\\
D\textsuperscript{2}L\textsuperscript{2}R\textsuperscript{2}~\cite{D2L2R2} & 66.10 & 73.20 \Tstrut\Bstrut\\
FDDL~\cite{FDDL}         & 65.22 & 73.64 \Tstrut\Bstrut\\
Our method \textbf{without} structural incoherence & 66.02 & 73.71 \Tstrut\Bstrut\\
Our method \textbf{with} structural incoherence    & 71.97 & 79.87 \Tstrut\Bstrut\\
\hline
\end{tabular}
\end{table}
%-------------------------------------------------------------------------------------------------------
%&&&&&&&&&&&&&&&&&&&&&&&&&&&&&&&&&&&&&&&&&&&&&&&&&&&&&&&&&&&&&&&&&&&&&&&&&&&&&&&&&&&&&&&&&&&&&&&&&&&&&&&&

In this experiment, to demonstrate the effect of structural incoherence term, we evaluate the recognition rate of JP-LRDL with and without this term. According to the results, our method with structural incoherence term, is superior to other approaches. Incorporating the structural incoherence term, would noticeably enhance the recognition rate, especially in datasets like the Caltech, that has large intra-class variations. Similarly, Figure~\ref{fig:Rep_Error} also verifies the role of structural incoherence term by presenting the representation error, with and without this term on a subset of the Caltech-101 dataset. The combination of LR and incoherence constraints helps us obtain a better estimate of the underlying distribution of samples and learn a robust and discriminative subspace. As a result, JP-LRDL is able to recognize objects in images despite imaging variations such as scale, viewpoint, lighting and background. 

To verify the robustness of our method to small-sized datasets, we select different numbers of training sample and train it on $\left \{5, 10, 15, 20, 25, 30\right \}$ images per category, and test on the rest. To compensate for the variation of the class size, we normalize the recognition results by the number of test images to get per-class accuracies. The final recognition accuracy is then obtained by averaging per-class accuracies across $102$ categories. We also repeat this experiment, by replacing a randomly located block of each test image with an unrelated image, such that 20\% pixels of every test image are occluded. The recognition rates are reported in Figures~\ref{fig:Caltech-Org-Train-Size} and~\ref{fig:Caltech-Noise-Train-Size} for the original and occluded images, respectively. Thanks to the efficiency of the proposed JP-LRDL, our method is able to achieve superior recognition rate, even when the number of training samples is relatively low. Although, the existing methods fail in occluded scenario, the proposed JP-LRDL still maintains satisfactory performance.
%-------------------------------------------------------------------------------------------------------
\subsection{Comparison to Deep Learning}
Learning through deep neural networks has recently drawn significant attention especially for image classification, and one key ingredient for this success is the use of convolutional architectures. Many variations of CNNs have been proposed and demonstrated superior performance over existing shallow methods, in several challenging vision tasks. However, as we will see, such a architecture does not generalize so well to recognition tasks where target dataset is small-sized and diverse in content compared to the base dataset, and has large intra-class variability.

Since we could not find any work that successfully applies CNN to the same recognition tasks, we use Caffe framework~\cite{Caffe} and select a pre-trained network on large-scale ImageNet dataset and then fine-tune it using the target data set for $1000$ epochs. We select two popular architectures; AlexNet~\cite{Alex-Net} and VGGNet-D~\cite{VGG-Net}, which their architecture consist of convolutional filter bank layers, nonlinear processing layers, and feature pooling layers. We evaluate these networks on five target datasets, each of which are known for different kind of intra-class variation, including illumination/viewpoint/pose changes, occlusion, disguise, background, etc. To challenge these architectures, we also simulate various levels of corruption and occlusion in the target datasets. For all the datasets, we follow the similar experiment protocol (\eg number of train and test samples), as already been described. The evaluation results are given in Table~\ref{table:Deep}. Here, $n_i$ shows the number of training samples per class to construct the target dataset.

We observe that these deep architectures do not perform well for none of the face-related experiments and this becomes worse, when there is simulated corruption or occlusion. The reason could be two things; the target dataset is smaller in size, but very different in content compared to the original dataset. Recent research~\cite{Transferable} reveal that complex models like CNNs is prone to overfitting when the target data is relatively small, and also the effectiveness of feature transfer is declined when the base and target tasks become less similar. In ImageNet, any kind of human face, with very large intra-class variation is categorized as "person and individual" class; however, the target task is face recognition, in which, unique individual should be classified as one class, despite of pose, expression and illumination changes and occlusion. We notice that, for object recognition task such as the Caltech-101 dataset, which has more training samples and more similar content to ImageNet, compared to face datasets, CNNs outperform JP-LRDL; however, when the data are corrupted by simulated noise, their performance drop significantly. One may say, CNNs are not the best model for classification of small-sized datasets with large intra-class variation, especially when the base and target datasets are different in content. It has not escaped our notice that, there could be some deep networks that may have great classification performance for these datasets; but, finding the best architecture for these datasets is not a trivial task and the learning critically depends on expertise of parameter tuning and some ad hoc tricks.
%&&&&&&&&&&&&&&&&&&&&&&&&&&&&&&&&&&&&&&&&&&&&&&&&&&&&&&&&&&&&&&&&&&&&&&&&&&&&&&&&&&&&&&&&&&&&&&&&&&&&&&&&
% --------------------------- Deep networks recognition results -----------------------------------------
\begin{table}[t]
\caption{Recognition rates (\%) of deep networks on different scenarios of various datasets}
\label{table:Deep}
\centering
\begin{tabular}{|l|c|l|p{0.9cm}|p{0.9cm}|p{1.1cm}|}
\hline
Dataset & Extra Challenge & $n_i$ & AlexNet & VGGNet & JP-LRDL \Tstrut\Bstrut\\
\hline \hline
Ext. YaleB & -----                  & 20 & 43.20 &  60.54 & 94.61 \Tstrut\\  
Ext. YaleB & 20\% corruption        & 20 & 27.41 &  41.31 & 88.01 \Tstrut\\
Ext. YaleB & 60\% corruption        & 20 & 16.40 &  23.65 & 54.31 \Tstrut\\
Ext. YaleB & 20\% occlusion         & 20 & 25.43 &  40.54 & 89.30 \Tstrut\\
Ext. YaleB & 60\% occlusion         & 20 & 14.51 &  22.54 & 64.42 \Tstrut\Bstrut\\
\hline
AR-Sunglasses  & -----            & 8 & 30.33 &  45.10 & 95.31 \Tstrut\\
AR-Sunglasses  & 20\% corruption  & 8 & 15.53 &  30.24 & 92.85 \Tstrut\\
AR-Scarf  & -----                 & 8 & 30.12 &  43.90 & 94.69 \Tstrut\\
AR-Scarf  & 20\% corruption       & 8 & 13.04 &  27.02 & 92.00 \Tstrut\\
AR-Mixed  & -----                 & 9 & 30.17 &  43.33 & 95.12 \Tstrut\Bstrut\\
AR-Mixed  & 20\% corruption       & 9 & 14.55 &  28.10 & 93.00 \Tstrut\Bstrut\\
\hline
LFWa & -----                      & 10 & 40.31 &  57.22 & 79.87 \Tstrut\Bstrut\\  
\hline
COIL-20 & -----                         & 10 & 65.70 & 70.76 & 92.10 \Tstrut\\
COIL-20 & 30\% corruption               & 10 & 19.35 & 36.45 & 56.72 \Tstrut\\
COIL-20 & 30\% occlusion                & 10 & 17.42 & 34.98 & 54.33 \Tstrut\Bstrut\\
\hline
Caltech-101 & -----                     & 30 & 81.15 & 89.10 & 79.87 \Tstrut\\
Caltech-101 & 20\% occlusion            & 30 & 65.20 & 69.12 & 71.00 \Tstrut\Bstrut\\
\hline
\end{tabular}
\end{table}
%-------------------------------------------------------------------------------------------------------
%&&&&&&&&&&&&&&&&&&&&&&&&&&&&&&&&&&&&&&&&&&&&&&&&&&&&&&&&&&&&&&&&&&&&&&&&&&&&&&&&&&&&&&&&&&&&&&&&&&&&&&&&
%-------------------------------------------------------------------------------------------------------
\section{Conclusion}
\label{sec:conclusion}
In this paper, we proposed an object classification method for small-sized datasets, which have large intra-class variation. The proposed method simultaneously learns a robust projection and a discriminative dictionary in the low-dimensional space, by incorporating LR, structural incoherence and dual graph constraints. These constraints would enables us to handle different types of intra-class variability, arising from different lightings, viewpoint and pose changes, occlusion and corruptions, even when there are few training samples per class. In the proposed joint DR and DL framework, learning can be performed in the reduced dimensions with lower computational complexity. Besides, by promoting the discriminative ability of the learned projection and dictionary, the projected samples can better preserve the discriminative information in relatively low dimensions; hence, JP-LRDL has superior performance even with a few number of features. Experimental results on different benchmark datasets validated the superior performance of JP-LRDL on image classification task especially when those few training samples are occluded, corrupted or captured under different lighting and viewpoint conditions.
%-------------------------------------------------------------------------------------------------------
\bibliographystyle{IEEEtran}
\bibliography{Homa-Arxiv}
\end{document}